\def\paperlanguage{}
\pdfoutput=1 % for arxiv

\documentclass[letterpaper, 10 pt, conference]{ieeeconf}  % Comment this line out if you need a4paper

\usepackage{bm}
\usepackage{cite}
\usepackage{flushend}
\include{preamble}

\newcommand{\argmin}{\mathop\textrm{arg~min}\limits}
\newcommand{\defeq}{\mathrel{\mathop:}=}

\IEEEoverridecommandlockouts                              % This command is only needed if
\title{\LARGE \textbf
  {
    \switchlanguage%
    {%
      Stable Tool-Use with Flexible Musculoskeletal Hands\\by Learning the Predictive Model of Sensor State Transition
    }%
    {%
      センサ状態遷移の予測モデル学習に基づく柔軟筋骨格ハンドの道具把持安定化戦略
    }%
  }
}

\author{Kento Kawaharazuka$^{1}$, Kei Tsuzuki$^{1}$, Moritaka Onitsuka$^{1}$, Yuki Asano$^{1}$\\Kei Okada$^{1}$, Koji Kawasaki$^{2}$,  and Masayuki Inaba$^{1}$% <-this % stops a space
  \thanks{$^{1}$ The authors are with the Department of Mechano-Informatics, Graduate School of Information Science and Technology, The University of Tokyo, 7-3-1 Hongo, Bunkyo-ku, Tokyo, 113-8656, Japan.
    {\texttt\small [kawaharazuka, tsuzuki, onitsuka, asano, k-okada, inaba]@jsk.t.u-tokyo.ac.jp}
  }
  \thanks{$^{2}$ The author is associated with TOYOTA MOTOR CORPORATION.
    {\texttt\small koji\_kawasaki@mail.toyota.co.jp}
  }
}
\begin{document}

\maketitle
\thispagestyle{empty}
\pagestyle{empty}

%%%%%%%%%%%%%%%%%%%%%%%%%%%%%%%%%%%%%%%%%%%%%%%%%%%%%%%%%%%%%%%%%%%%%%%%%%%%%%%%
\begin{abstract}
  \switchlanguage%
  {%
    The flexible under-actuated musculoskeletal hand is superior in its adaptability and impact resistance.
    On the other hand, since the relationship between sensors and actuators cannot be uniquely determined, almost all its controls are based on feedforward controls.
    When grasping and using a tool, the contact state of the hand gradually changes due to the inertia of the tool or impact of action, and the initial contact state is hardly kept.
    In this study, we propose a system that trains the predictive network of sensor state transition using the actual robot sensor information, and keeps the initial contact state by a feedback control using the network.
    We conduct experiments of hammer hitting, vacuuming, and brooming, and verify the effectiveness of this study.
  }%
  {%
    柔軟で劣駆動な筋骨格ハンドはその適応性や耐衝撃性において優れいている.
    一方, センサとアクチュエータの関係を一意に定めることができず, そのほとんどがフィードフォワードな制御に基づいている.
    そのため, 道具を把持して用いる場合, 動作中の道具の慣性や衝撃から徐々に接触状態が変化し, 初期の接触状態を常に保ち続けられるとは限らない.
    そこで本研究では, 接触状態遷移の予測モデルを実機データから学習させ, それを元にフィードバックを行うことで初期接触状態を常に保ち続ける手法を提案する.
    ハンマー, 箒, 掃除機の操作に関して実験を行い, 本研究の有効性を確認した.
  }%
\end{abstract}

\section{INTRODUCTION}\label{sec:introduction}
\begin{figure}[t]
  \centering
  \includegraphics[width=1.0\columnwidth]{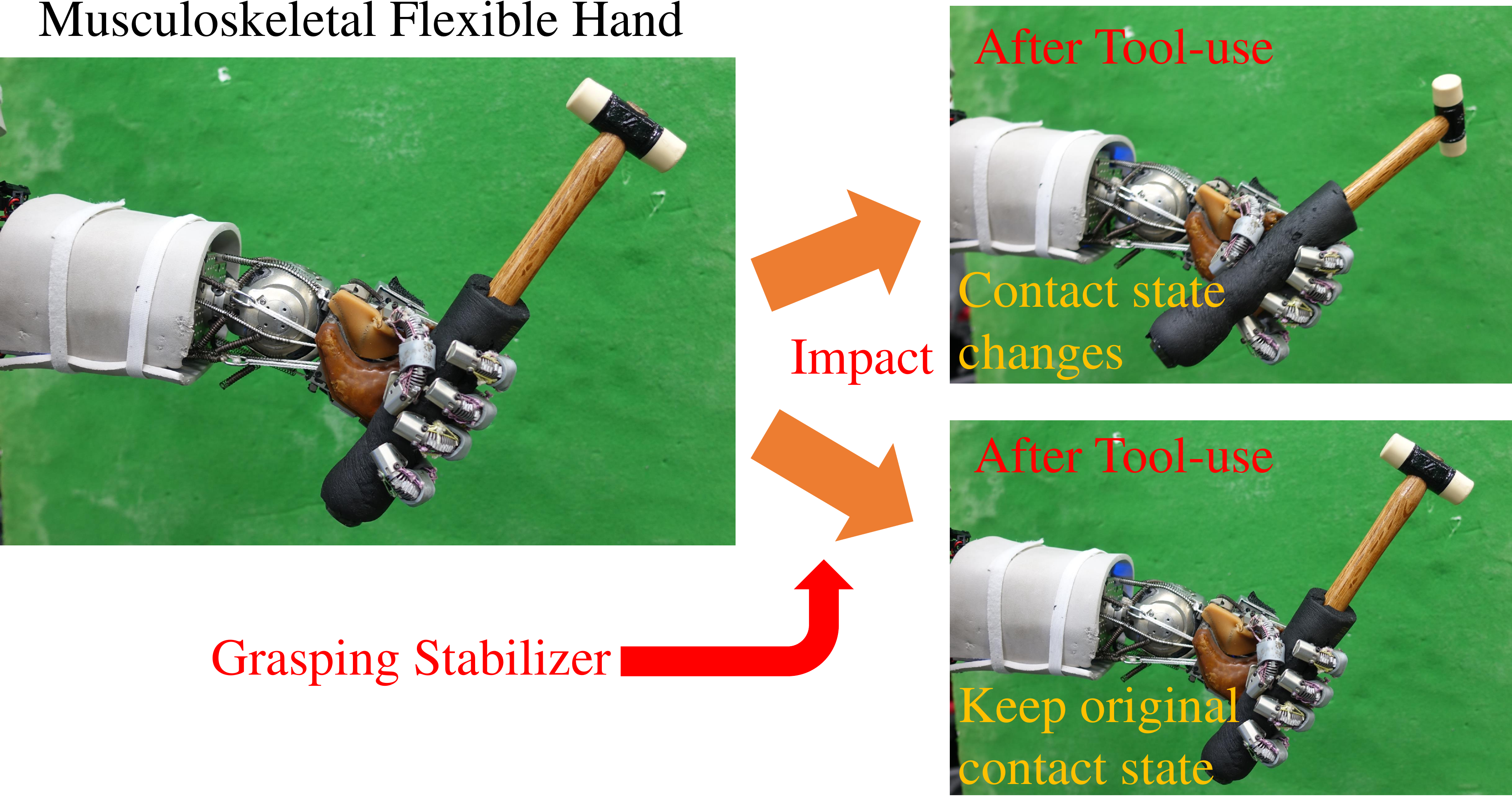}
  \caption{Grasping stabilizer for tool-use.}
  \label{figure:motivation}
  \vspace{-3.0ex}
\end{figure}

\switchlanguage%
{%
  Various robotics hands \cite{kochan2005shadowhand, grebenstein2011dlrhand, kim2014roborayhand, deimel2016underactuatedhand, wiste2017anthrohand, xu2016biohand, kontoudis2015lockablehand, makino2017hand, makino2018hand} have been developed so far.
  While many hands with dozens of tendons for dexterous manipulation \cite{kochan2005shadowhand, grebenstein2011dlrhand, kim2014roborayhand} exist, soft robotic hands such as the flexible pneumatic hands \cite{deimel2016underactuatedhand} and tendon-driven under-actuated hands \cite{wiste2017anthrohand, xu2016biohand, kontoudis2015lockablehand, makino2017hand, makino2018hand} have prevailed thanks to the recent growth of soft robotics \cite{lee2017softrobotics}.
  These hands have few actuators and are usually under-actuated, and its joints or links are often composed of rubber or springs.
  They can grasp objects adaptively thanks to the flexibility even with few actuators and are superior in impact resistance.
  Several of them can exert high grip force with a few strong actuators \cite{makino2017hand, makino2018hand}.

  Feedforward controls such as applying constant force or keeping a constant grasp shape are usually used for these flexible hands.
  It is because feedback controls are challenging, since the modelization of under-actuated flexible hands with soft joints and links is difficult, and the relationship between sensors and actuators cannot be uniquely determined.
  When grasping and using a tool, as shown in \figref{figure:motivation}, the contact state gradually changes due to the inertia of the tool or impact of action, and the initial contact state is hardly kept.
  The robot sometimes drops the grasped tool or the posture of the grasped tool can change.

  To solve the problem, regarding fully-actuated hands and simple under-actuated hands, real-time tactile feedback and regrasp planning have been developed by limiting manipulation plane and using accurate modelings of kinematics and dynamics.
  In \cite{allen1997feedback}, a real-time change in grasping behavior on a 2D plane is implemented using vision and strain gauge.
  In \cite{bicchi1989feedback}, regarding predefined grasp shapes, force optimization to stabilize the grasping using tactile sensors is discussed.
  In \cite{regoli2016grasp}, a force feedback using three fingers on a 2D plane is implemented.
  In \cite{schmid2008door}, the robot can open the door accurately by using tactile sensor information.
  In \cite{hogan2018regrasp, calandra2018regrasp}, a regrasp planning to grasp the object more stably is realized.
  In \cite{li2015garments}, accurate cloth folding is realized by judging the success of grasping using vision and by regrasp planning.

  On the other hand, thanks to the recent growth of deep learning, various learning-based methods have been developed.
  In \cite{chebotar2016regrasping}, a regrasp planning with reinforcement learning is realized by predicting the success of grasping.
  In \cite{jain2019manipulation}, an imitation learning based method to train deep visuomotor policies for various manipulation tasks with a simulated five fingered dexterous hand is developed.
  In \cite{hoof2015reinforcement}, regarding an under-actuated robot hand, in-hand manipulation on a 2D plane using two fingers is realized using reinforcement learning.
  In \cite{homberg2019soft}, a classification of grasped objects is realized using a pneumatic flexible hand.
  However, many studies with reinforcement learning are conducted only in simulation.
  Regarding flexible hands, because their simulation is difficult, almost all studies are about classification of grasped objects and regrasp planning.
  There exist few studies about in-hand manipulation or real-time tactile feedback, and they have not focused on stable tool-use by flexible hands.
}%
{%
  これまで多くのロボットハンドが開発されてきた\cite{kochan2005shadowhand, grebenstein2011dlrhand, kim2014roborayhand, deimel2016underactuatedhand, wiste2017anthrohand, xu2016biohand, kontoudis2015lockablehand, makino2017hand, makino2018hand}.
  数十本にも及ぶ腱により精巧なマニピュレーションを可能とするハンド\cite{kochan2005shadowhand, grebenstein2011dlrhand, kim2014roborayhand}が多く存在する一方, ソフトロボティクスの台頭\cite{lee2017softrobotics}により, 空気圧により駆動する柔軟ハンド\cite{deimel2016underactuatedhand}, 劣駆動な指を持つ腱駆動柔軟ハンド\cite{wiste2017anthrohand, xu2016biohand, kontoudis2015lockablehand, makino2017hand, makino2018hand}が普及し始めている.
  これらは指が劣駆動でアクチュエータ数を絞り, 関節はゴムやバネ等によって構成されている場合が多い.
  少ないアクチュエータでもその柔軟性により適応的な把持が可能であり, 耐衝撃性に優れ, 少数の大きなアクチュエータを用いることで高把持力を実現することもできる\cite{makino2017hand, makino2018hand}.

  これらの柔軟ハンドは基本的にはフィードフォワードに制御入力を決め, 一定の把持形態を維持したり, 一定の力を加え続けることが多い.
  劣駆動で, 関節やリンクの組織自体が柔軟なハンドのモデル化は非常に難しく, センサとアクチュエータの関係を一意に決めることができなく, フィードバック制御が困難なためである.
  そのため, 例えば道具を把持して用いる場合, \figref{figure:motivation}のように, 動作中に道具の慣性や衝撃から徐々に接触状態が変化し, 初期の接触状態を常に保ち続けられるとは限らない.
  接触状態の変化によって, ときには把持物体を落としたり, 把持した道具の姿勢が変化してしまうことがしばしばある.

  一方, 全駆動ハンドやモデル化容易な劣駆動ハンドでは, 操作平面の限定や動作の制限を含めたうえで, 正確なモデル化と視覚・接触センサの利用によって, リアルタイムのtactile feedbackや再把持計画が行われている.
  \cite{allen1997feedback}では視覚とひずみゲージを用いて, リアルタイムに2次元平面上での動作変更を行っている.
  \cite{bicchi1989feedback}では, 決まった把持形態に関して把持を安定化させるための接触センサを用いた力最適化について議論している.
  \cite{regoli2016grasp}では, 手のひらは考えず, 3指の接触による平面上の動きのみ考えて安定した把持実現のための力フィードバックを行っている.
  \cite{schmid2008door}では, 接触センサ情報から正確にドアを開ける動作を実現している.
  \cite{hogan2018regrasp, calandra2018regrasp}では, 一度把持した際の接触センサ値から, より安定した接触を実現できるよう再把持を行う.
  \cite{li2015garments}は視覚により把持が成功したかどうかを判定し, ダメなら再把持をすることで, しっかりと衣服を広げてたたむ

  これらに対して, 深層学習の台頭により, 様々な学習ベースの手法が開発されてきている.
  \cite{chebotar2016regrasping}は把持成功率の予測と強化学習による再把持を実現している.
  \cite{jain2019manipulation}はimitation learning based approach to train deep visuomotor policies for a variety of manipulation tasks with a simulated five fingered dexterous handしている.
  \cite{hoof2015reinforcement}は劣駆動なロボットハンドに対して接触センサを用いて強化学習を適用し, 二次元平面上で二指によるIn-hand manipulationを実行している.
  \cite{homberg2019soft}は空気圧駆動の劣駆動柔軟ハンドによる把持物体の分類を行っている.
  しかし, 強化学習の多くはシミュレーションであり, 柔軟ハンドのシミュレーションは難しいため, 把持物体の分類等が多く, In-hand manipulationや触覚フィードバックを実現した研究は少ない.
  また, 道具使用の際の把持安定化に着目した研究はない.
}%

\begin{figure*}[t]
  \centering
  \includegraphics[width=1.9\columnwidth]{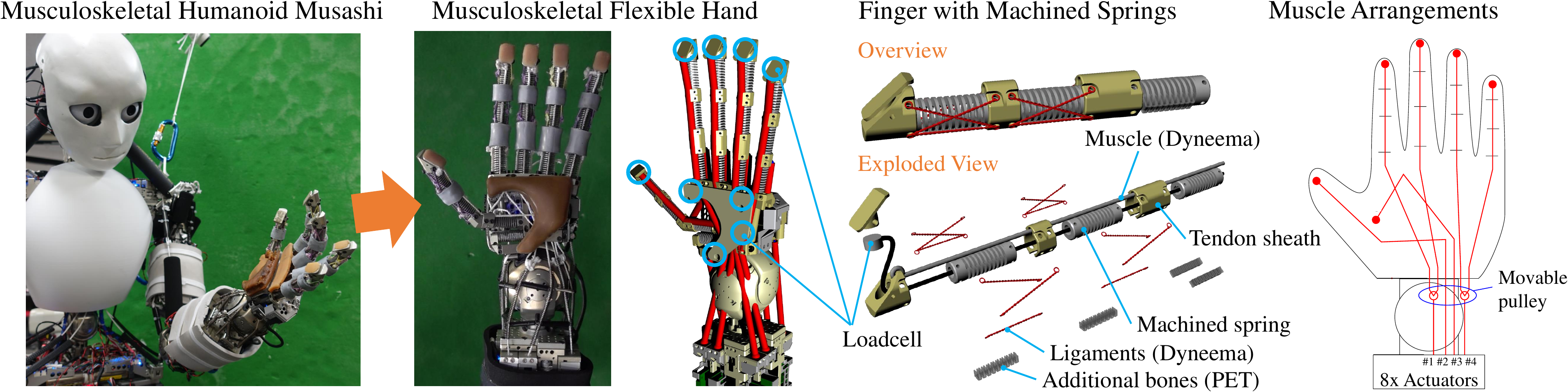}
  \caption{Five-fingered musculoskeletal flexible hand \cite{makino2018hand} installed in the musculoskeletal humanoid Musashi \cite{kawaharazuka2019musashi}.}
  \label{figure:musculoskeletal-hand}
  \vspace{-3.0ex}
\end{figure*}

\switchlanguage%
{%
  In this study, for stable tool-use by flexible hands, we propose a feedback control to keep the initial contact state by training the predictive model of sensor state transition expressed by a neural network.
  On the basis of previous studies \cite{kawaharazuka2019dynamic, kawaharazuka2019pedal}, we explore random search behavior, loss function, and optimization method, and propose a novel grasping stabilizer focusing on stable tool-use.
  We apply this study to the five-fingered musculoskeletal hand installed in the musculoskeletal humanoid Musashi \cite{kawaharazuka2019musashi}, and verify the effectiveness of this study by experiments of hammer hitting, vacuuming, and brooming.

  % In \secref{sec:musculoskeletal-hand}, we will explain the musculoskeletal flexible hand \cite{makino2018hand}.
  % In \secref{sec:proposed-method}, we will propose a grasping stabilizer based on the construction of a predictive model of sensor state transition.
  % In \secref{sec:experiments}, we will explain results of model training, prediction accuracy, and stabilization, regarding concrete tasks.
  % Finally, we will state the discussion and conclusion about the experiments and methods.
}%
{%
  そこで本研究では, 柔軟ハンドによる道具把持の際の, 接触状態遷移予測モデル構築とそれを用いたフィードバック制御による接触状態維持フィードバック制御を行う.
  \cite{kawaharazuka2019dynamic, kawaharazuka2019pedal}をベースとして探索や損失関数, 最適化等を新たにし, 道具把持に着目した新しい把持安定化制御を提案する.
  筋骨格ヒューマノイドMusashi \cite{kawaharazuka2019musashi}に搭載された筋骨格五指ハンド\cite{makino2018hand}に本手法を適用し, ハンマー打撃動作, 箒掃き動作, 掃除動作を行うことで, 本研究の有効性を確認する.

  第二章では, 本研究で使用する筋骨格柔軟ハンド\cite{makino2018hand}について説明する.
  第三章では, 接触状態遷移の予測モデル構築に基づく把持安定化戦略について述べる.
  第四章では, 具体的なタスクにおいて, 実際のモデルの学習, 予測精度, 把持安定化等の結果について詳細に述べる.
  最後に, それら手法と実験について議論と結論を述べる.
}%

\section{Musculoskeletal Flexible Hand} \label{sec:musculoskeletal-hand}
\switchlanguage%
{%
  As shown in \figref{figure:musculoskeletal-hand}, the musculoskeletal flexible hand \cite{makino2018hand} installed in the musculoskeletal humanoid Musashi \cite{kawaharazuka2019musashi} has five fingers, and each finger is composed of three flexible machined springs.
  PET plates and strings imitating ligaments are attached to the machined springs to make anisotropy in fingers.
  Dyneema is arranged around the machined spring as a muscle.

  Eight muscle actuators \cite{kawaharazuka2017forearm} are equipped in the forearm of Musashi; three of them are for movements of the wrist, and five of them are for fingers.
  Two of the five finger muscles actuate index/middle fingers and ring/little fingers using a movable pulley.
  The Other two of the five finger muscles actuate the thumb.
  Also, the remaining finger muscle can change the stiffness of the fingers by raising muscle tension and compressing machined springs.
  Among these five muscles, we choose four muscles directly involved with movements of fingers (the former four) as control input.
  We represent target muscle lengths as $\bm{l}^{target}$ and measured muscle lengths from encoders as $\bm{l}$.

  Nine loadcells as contact sensors are equipped in each finger tip and the palm, and their arrangement is shown in the middle figure of \figref{figure:musculoskeletal-hand}.
  Also, muscle tensions are measured from muscle actuators \cite{kawaharazuka2017forearm}.
  We represent the loadcell values as $\bm{C}$ and muscle tension values as $\bm{F}$.

  Thus, in this study, $\bm{F}$ and $\bm{C}$ are controlled by $\bm{l}^{target}$.
  $\bm{l}$ and $\bm{F}$ are four dimensional vectors, and $\bm{C}$ is a nine dimensional vector.
}%
{%
  本研究で使用する筋骨格柔軟ハンド\cite{makino2018hand}について説明し, そのセンサやアクチュエーションについて触れる.
  \figref{figure:musculoskeletal-hand}に示すように, 筋骨格ヒューマノイドMusashi \cite{kawaharazuka2019musashi}に搭載されたこのハンドは五指を有し, それぞれの指の関節は柔軟な切削バネによって構成されている.
  切削ばねは異方性を出すために靭帯を模したDyneemaとPETプレートが付属している.
  一つの指は3つの切削バネからなり, その中にロードセルのケーブルが, その外に筋としてのDyneemaが存在している.

  Musashi \cite{kawaharazuka2019musashi}の前腕には, 8本の筋アクチュエータ\cite{kawaharazuka2017forearm}が存在し, 手首に3本, 指に5本が割り当てられている. 
  指の5本の腱のうち2本はそれぞれ人差し指と中指, 薬指と小指を駆動しており, 2つの指をプーリで分岐することで制御している.
  また, うち2本は親指を駆動している.
  最後の一本は張力を高めることで切削ばねを押しつぶし, 指の剛性を変化させることができる.
  この5本のうち, 直接指に関わる4本のアクチュエータを制御対象とし, その制御指令値を$\bm{l}^{target}$, エンコーダから得られる現在筋長を$\bm{l}$とする.

  それぞれの指の先端, 手のひらには9つの接触センサとしてのロードセルが分布しており, その配置は\figref{figure:musculoskeletal-hand}の中図のようになっている.
  また, 筋アクチュエータ\cite{kawaharazuka2017forearm}からは筋張力を測定することができる.
  このロードセル値を$\bm{C}$, 筋張力値を$\bm{F}$とする.

  つまり本研究では, $\bm{l}^{target}$をもとに, $\bm{F}$, $\bm{C}$を制御することになる.
  $\bm{l}$は4次元, $\bm{F}$は4次元, $\bm{C}$は9次元である.
}%

\section{Grasping Stabilizer} \label{sec:proposed-method}
\switchlanguage%
{%
  The overall procedures of this study are as below,
  \begin{enumerate}
    \item Random search behavior of tool grasping by random control input
    \item Training of the predictive model of sensor state transition
    \item Stable tool-use by grasping stabilizer using the predictive model
  \end{enumerate}
}%
{%
  本研究全体の手順は以下のようになっている.
  \begin{enumerate}
    \item 道具把持状態における制御入力探索動作
    \item 接触状態遷移の予測モデル学習
    \item 予測モデルを用いた道具使用動作における把持安定化
  \end{enumerate}
}%

\subsection{Formulation of This Study} \label{subsec:formulation}
\switchlanguage%
{%
  We formulate the problem handled in this study.
  We represent the contact state with muscle tensions $\bm{F}$ and loadcell values $\bm{C}$ as $\bm{s} = (\bm{F}^T, \bm{C}^{T})^{T}$.
  Also, we represent target muscle lengths $\bm{l}^{target}$ as control input $\bm{u}$, muscle lengths $\bm{l}$ and muscle length velocities $\dot{\bm{l}}$ as control state $\bm{i}=(\bm{l}, \dot{\bm{l}})^{T}$.
  The predictive model of sensor state transition is formulated as below,
  \begin{align}
    \bm{s}_{[t+1, t+T]} &\defeq (\bm{s}^{T}_{t+1}, \bm{s}^{T}_{t+2}, \cdots, \bm{s}^{T}_{t+T})^{T}\nonumber\\
    \bm{u}_{[t, t+T-1]} &\defeq (\bm{u}^{T}_{t}, \bm{u}^{T}_{t+1}, \cdots, \bm{u}^{T}_{t+T-1})^{T}\nonumber\\
    \bm{s}_{[t+1, t+T]} &= \bm{f}((\bm{s}^{T}_{t}, \bm{i}^{T}_{t}, \bm{u}^{T}_{[t, t+T-1]})^{T})\label{eq:model}
  \end{align}
  where $\bm{f}$ is the predictive model, $t$ expresses the current timestep, and $T$ expresses how many timesteps ahead to predict.

  $\bm{f}$ is trained using the actual robot sensor information.
  After that, we conduct the grasping stabilizer using this trained $\bm{f}$.
  We represent the initial contact state, when grasping the tool by a feedforward control, as $\bm{s}^{keep}$.
  The control input $\bm{u}^{opt}_{[t, t+T-1]}$ to keep $\bm{s}^{keep}$ is calculated as below,
  \begin{align}
    \bm{s}^{predict}_{seq} &\defeq \bm{f}((\bm{s}^{T}_{t}, \bm{i}^{T}_{t}, (\bm{u}^{init}_{seq})^{T})^{T})\nonumber\\
    \bm{u}^{opt}_{seq} &\defeq \argmin_{\bm{u}^{min}\leq\bm{u}^{init}\leq\bm{u}^{max}} L_{opt}(\bm{s}^{predict}_{seq}, \bm{s}^{keep}_{seq}, \bm{u}^{init}_{seq})\label{eq:opt}
  \end{align}
  where $\bm{u}^{\{min, max\}}$ is the minimum or maximum value of $\bm{u}$, $L_{opt}$ is a loss function for optimization, $\bm{s}^{\{predict, keep\}}_{seq}$ is an abbreviation of $\bm{s}^{\{predict, keep\}}_{[t+1, t+T]}$, $\bm{u}^{\{init, keep\}}_{seq}$ is an abbreviation of $\bm{u}^{\{init, keep\}}_{[t, t+T-1]}$, and $\bm{s}^{keep}_{[t+1, t+T]}$ is the vector which duplicates $T$ number of $\bm{s}^{keep}$.
  Since $\bm{s}_{[t+1, t+T]}$ cannot be obtained at the timestep $t$, the contact state $\bm{s}^{predict}_{seq}$ predicted by $\bm{f}$ and initial control input $\bm{u}^{opt}_{seq}$ before optimization is used for this value.
  $L_{opt}$ must be calculated to make close $\bm{s}^{predict}_{seq}$ and $\bm{s}^{keep}_{seq}$, and to output an executable $\bm{u}^{opt}_{seq}$.
  \equref{eq:opt} is conducted to calculate $\bm{u}^{opt}_{[t, t+T-1]}$ at every timestep.

  It takes much time to calculate \equref{eq:opt}, and $\bm{u}^{opt}_{[t, t+T-1]}$ is calculated using the interval from the current timestep to the next timestep.
  Thus, not $\bm{u}^{opt}_{t}$ but $\bm{u}^{opt}_{t+1}$ is sent to the actual robot.
}%
{%
  本研究で扱う問題について定式化を行う.
  本研究では筋張力$\bm{F}$とロードセルの接触センサ値$\bm{C}$を接触状態$\bm{s} = (\bm{F}^T, \bm{C}^{T})^{T}$として扱う.
  また, 指令筋長$\bm{l}^{target}$を制御入力$\bm{u}$, 筋長$\bm{l}$と筋長速度$\dot{\bm{l}}$を制御状態$\bm{i}=(\bm{l}, \dot{\bm{l}})^{T}$とする.
  本研究における接触状態遷移の予測モデルは, 以下の式で表される.
  \begin{align}
    \bm{s}_{[t+1, t+T]} &\defeq (\bm{s}^{T}_{t+1}, \bm{s}^{T}_{t+2}, \cdots, \bm{s}^{T}_{t+T})^{T}\nonumber\\
    \bm{u}_{[t, t+T-1]} &\defeq (\bm{u}^{T}_{t}, \bm{u}^{T}_{t+1}, \cdots, \bm{u}^{T}_{t+T-1})^{T}\nonumber\\
    \bm{s}_{[t+1, t+T]} &= \bm{f}((\bm{s}^{T}_{t}, \bm{i}^{T}_{t}, \bm{u}^{T}_{[t, t+T-1]})^{T})\label{eq:model}
  \end{align}
  ここで, $\bm{f}$は予測モデル, $t$は現在のタイムステップ, $T$は何ステップ先まで予測するかを表す.

  この$\bm{f}$を実機データを用いて学習する.
  その後, この$\bm{f}$を用いて把持安定化を行う.
  最初にフィードフォワード的に掴んだ際の接触状態を$\bm{s}^{keep}$とする.
  この$\bm{s}^{keep}$を保ち続ける制御入力$\bm{u}^{opt}_{[t, t+T-1]}$を以下のように求める.
  \begin{align}
    \bm{s}^{predict}_{seq} &\defeq \bm{f}((\bm{s}^{T}_{t}, \bm{i}^{T}_{t}, (\bm{u}^{init}_{seq})^{T})^{T})\nonumber\\
    \bm{u}^{opt}_{seq} &\defeq \argmin_{\bm{u}^{min}\leq\bm{u}^{init}\leq\bm{u}^{max}} L_{opt}(\bm{s}^{predict}_{seq}, \bm{s}^{keep}_{seq}, \bm{u}^{init}_{seq})\label{eq:opt}
  \end{align}
  ここで, $\bm{u}^{\{min, max\}}$は$\bm{u}$の最小値または最大値, $L_{opt}$は最適化のための損失関数, $\bm{s}^{\{predict, keep\}}_{seq}$は$\bm{s}^{\{predict, keep\}}_{[t+1, t+T]}$を省略したもの, $\bm{u}^{\{init, keep\}}_{seq}$は$\bm{u}^{\{init, keep\}}_{[t, t+T-1]}$を省略したもの, $\bm{s}^{keep}_{[t+1, t+T]}$は$\bm{s}^{keep}$を$T$個並べた値を表す.
  $\bm{s}_{[t+1, t+T]}$はタイムステップ$t$のときには分からないため, 制御入力の初期値$\bm{u}^{opt}_{seq}$を予測モデル$\bm{f}$に入力したときの予測値$\bm{s}^{predict}_{seq}$を用いる.
  この$\bm{s}^{predict}_{seq}$と$\bm{s}^{keep}_{seq}$の値が近くなり, かつ実行可能な$\bm{u}^{opt}_{seq}$が出力されるような損失関数$L_{opt}$を設計する必要がある.
  この\equref{eq:opt}を, 毎タイムステップ行い, $\bm{u}^{opt}_{[t, t+T-1]}$を計算し続ける.

  また, \equref{eq:opt}の計算には時間を要し, 現在のタイムステップから次のタイムステップまでの時間を使って$\bm{u}^{opt}_{[t, t+T-1]}$を求める.
  そのため, 実機に送る値は$\bm{u}^{opt}_{t}$ではなく, $\bm{u}^{opt}_{t+1}$となる.
}%

\subsection{Network Structure of Predictive Model} \label{subsec:basic-structure}
\switchlanguage%
{%
  As a network structure of \equref{eq:model}, we can consider various types.
  As one example, we can represent the network using LSTM \cite{hochreiter1997lstm}.
  $\bm{s}_{t+1}=\bm{f}_{rnn}((\bm{s}^{T}_{t}, \bm{u}^{T}_{t})^{T})$ is expressed by LSTM, and the network of \equref{eq:model} is constructed by extending the LSTM $T$ times.
  While this structure has benefits such as small model size and a changeable $T$, extending the model $T$ times successively takes much time and it is a disadvantage in calculating \equref{eq:opt}.

  In this study, by fixing $\bm{T}$, \equref{eq:model} is directly represented by a neural network with five fully connected layers including inputs and outputs.
  Because the network can directly calculate $\bm{s}_{[t+1, t+T]}$ with only one forwarding, it is an advantage for computational cost.
  The number of units in middle layers are set as $(100, 100, 100)$, and Batch Normalization \cite{ioffe2015batchnorm} and activation function Sigmoid are inserted in all layers except for the last layer.

  In this study, we set the unit of $\bm{l}$, $\bm{C}$, and $\bm{F}$ as [mm/10], [N/10], and [N/200], respectively, to align their average values.
  Also, the control frequency is 5 Hz because \equref{eq:opt} takes much time.
  Thus, \equref{eq:model} predicts the contact state until 2 sec ahead by setting $T=10$.
}%
{%
  \secref{subsec:formulation}で述べた予測モデル\equref{eq:model}について述べる.
  これには様々な実装方法が考えられる.
  一つとして, LSTM \cite{hochreiter1997lstm}を使った方法が考えられる.
  つまり, $\bm{s}_{t+1}=\bm{f}_{rnn}((\bm{s}^{T}_{t}, \bm{u}^{T}_{t})^{T})$をLSTMによって表現し, それを$T$回展開することで\equref{eq:model}と同じ形を得る方法である.
  これはモデルの小ささや展開数$\bm{T}$を可変にできるというメリットを有する一方で, モデルを連続的に展開するため計算に時間を要し, \equref{eq:opt}の計算の際に不利となる.

  本研究では, $\bm{T}$を固定して, 入力と出力を含めた5層のニューラルネットワークで直接\equref{eq:model}を表現する方法をとる.
  これは, 一度のフォーワードで$\bm{s}_{[t+1, t+T]}$を直接計算できるため, 計算速度の観点では有利となる.
  本研究では中間層のユニット数を$(100, 100, 100)$として, 最終層以外の全ての層にBatch Normalization \cite{ioffe2015batchnorm}と活性化関数Sigmoidを挿入している.

  本研究では値の大きさを揃えるために, $l$の単位を[mm/10], $C$の単位を[N/10], $F$の単位を[N/200]としている.
  また, 制御周期は速い必要はなく, \equref{eq:opt}の計算に時間を使うため5 Hzで行っており, $T=10$として2 sec先までの状態を\equref{eq:model}が予測する.
}%

\subsection{Random Search Behavior} \label{subsec:grasping-search}
\switchlanguage%
{%
  The actual robot sensor information is obtained to train the neural network stated in \secref{subsec:basic-structure}.
  First, when grasping a tool, a constant target muscle length $\bm{l}^{target}_{0}$ for the tool is sent feedforwardly.
  We define that $\bm{l}_{target}$ as control input $\bm{u}$ represents the difference from $\bm{l}^{target}_{0}$.
  Search behavior of tool grasping is shown below,
  \begin{align}
    \Delta\bm{u} &\defeq \bm{C}_{rand}\textrm{sin}(C_{time}{t})\\
    \bm{u} &= \bm{u} + \textrm{Random}(-\Delta\bm{u}, \Delta\bm{u})\\
    \bm{u} &= \textrm{max}(\bm{u}^{min}, \textrm{min}(\bm{u}, \bm{u}^{max}))
  \end{align}
  where $\bm{C}_{rand}$ and $C_{time}$ are coefficients to determine $\Delta\bm{u}$, and $\textrm{Random}(a, b)$ outputs a random value in the range of $[a, b]$.

  $\Delta\bm{u}$ can also be fixed at a constant value.
  However, one problem occurs in this case.
  In this study of handling stable tool-use, if impact or external force is not added, keeping $\bm{l}_{target}$ at a constant value is the best.
  When fixing $\Delta\bm{u}$ at a constant value, the data keeping $\bm{l}_{target}$ at a constant value cannot be obtained and the grasping stabilizer continues to break and return to the current contact state.
  Therefore, by changing $\Delta\bm{u}$ variably, various data can be obtained and the grasping stabilizer can work stably.

  By using these data, \equref{eq:model} is trained by setting the batch size as $C^{train}_{batch}$ and number of epochs as $C^{train}_{epoch}$.
  We use the loss function $L_{origin}$ when training, as shown below,
  \begin{align}
    L_{origin}(\bm{s}^{predict}_{seq}, \bm{s}^{keep}_{seq}) \defeq ||\bm{s}^{predict}_{seq}, \bm{s}^{keep}_{seq}||^{2}_{2} \label{eq:origin-loss}
  \end{align}
  where $||\cdot||_{2}$ expresses L2 norm, and so $L_{origin}$ is mean squared error loss.

  In this study, we set $\bm{u}^{min}=-5$ [mm], $\bm{u}^{max}=20$ [mm], $\bm{C}_{rand}=3$ [mm], $\bm{C}_{time}=0.02$ [1/timestep], $C^{train}_{batch}=10$, and $C^{train}_{epoch}=300$.
}%
{%
  \secref{subsec:basic-structure}で表現したモデルを学習させるために, 実機データを取得する.
  道具を握る際, まずフィードフォワード的にある一定の把持姿勢$\bm{l}^{target}_{0}$を送る.
  ここで, 制御入力$\bm{u}$となる$\bm{l}_{target}$は, $\bm{l}^{target}_{0}$からの差分とする.
  この際のランダムな制御則を以下のように定める.
  \begin{align}
    \Delta\bm{u} &\defeq \bm{C}_{rand}\textrm{sin}(C_{time}{t})\\
    \bm{u} &= \bm{u} + \textrm{Random}(-\Delta\bm{u}, \Delta\bm{u})\\
    \bm{u} &= \textrm{max}(\bm{u}^{min}, \textrm{min}(\bm{u}, \bm{u}^{max}))
  \end{align}
  ここで, $\bm{C}_{rand}$, $C_{time}$は$\Delta\bm{u}$を決める係数であり, $\textrm{Random}(a, b)$は$[a, b]$の範囲でランダムな値を出力する関数である.

  $\Delta\bm{u}$を一定の値に固定することも可能である.
  しかしこの場合問題が発生する.
  本研究のタスクにおいては, 衝撃や外力が加わらなければ, $\bm{l}_{target}=\bm{0}$を常に保ち続けることが最善である.
  そのため, $\Delta\bm{u}$を固定した場合だと, $\bm{l}_{target}=\bm{0}$を保つデータが得られれず, 常に接触状態を崩してそれを戻す動作を繰り返すような挙動が発生してしまう.
  よって, このように$\Delta\bm{u}$を変化させることで多様なデータを取得し, より安定して動作することが可能となる.

  このデータを用いて, \equref{eq:model}を, バッチサイズを$C^{train}_{batch}$, エポック数を$C^{train}_{epoch}$として学習させる.
  この際の損失関数は以下に示すような$L_{origin}$を用いる.
  \begin{align}
    L_{origin}(\bm{s}^{predict}_{seq}, \bm{s}^{keep}_{seq}) \defeq ||\bm{s}^{predict}_{seq}, \bm{s}^{keep}_{seq}||^{2}_{2} \label{eq:origin-loss}
  \end{align}
  ここで, $||\cdot||_{2}$はL2 normを表し, つまりこの$L_{origin}$は平均二乗誤差を意味する.

  本研究の実験では, $\bm{u}^{min}=-5$ [mm], $\bm{u}^{max}=20$ [mm], $\bm{C}_{rand}=3$ [mm], $\bm{C}_{time}=0.02$ [1/timestep], $C^{train}_{batch}=10$, $C^{train}_{epoch}$とする.
}%

\begin{figure}[t]
  \centering
  \includegraphics[width=1.0\columnwidth]{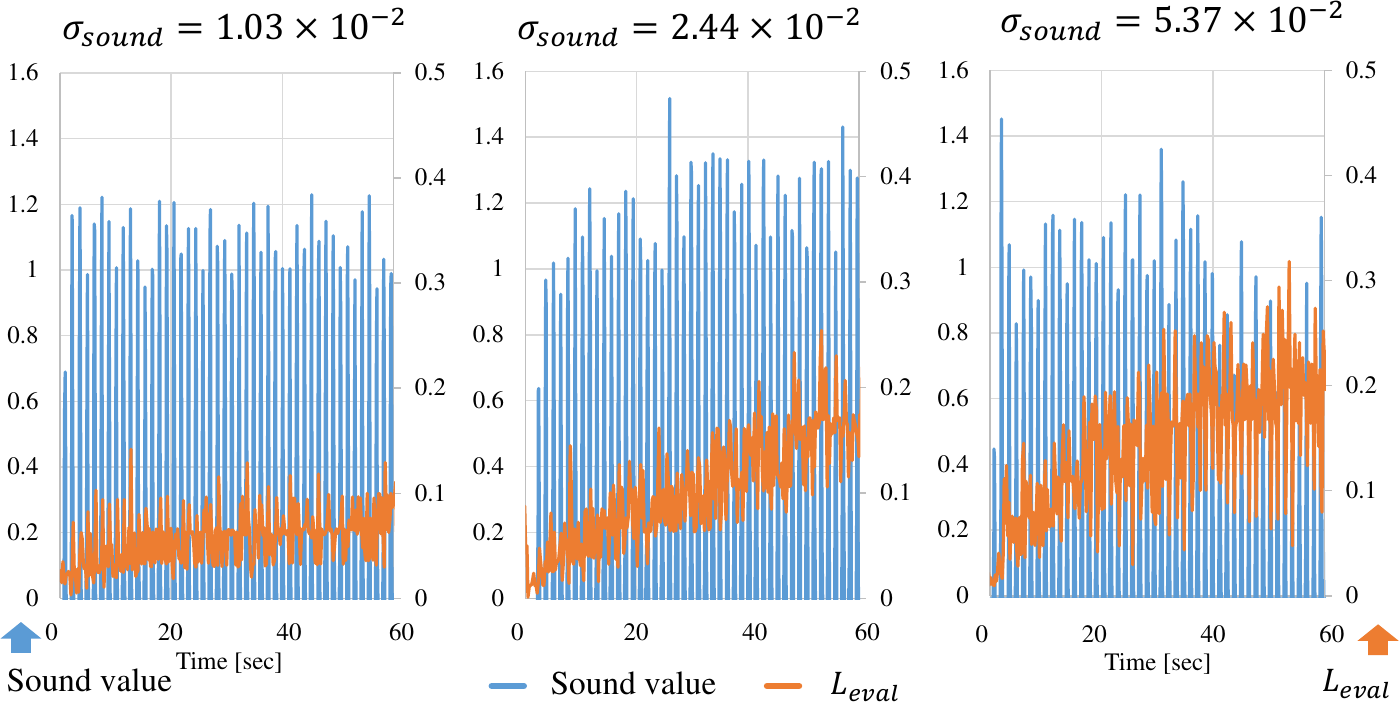}
  \caption{Experimental evaluation of the correlation between $L_{eval}$ and sound value when hitting a plate with a hammer.}
  \label{figure:loss-sound}
  \vspace{-3.0ex}
\end{figure}

\subsection{Loss Definition} \label{subsec:loss-definition}
\switchlanguage%
{%
  We will explain the design of $L_{opt}$ in \equref{eq:opt}.
  Here, we disregard the limitation of control input $\bm{u}$, and it will be discussed in \secref{subsec:grasping-stabilizer}.

  This study focuses on stabilizing tool-use, that is, keeping the initial contact state.
  We need to consider the characteristics of contact sensor and muscle tension sensor handled in this study.
  While these sensor values become $\bm{0}$ without any contact, the values can steadily increase until the rated values in the direction of strong contact.

  First, we assume that $L_{opt}$ is set as $L_{origin}$ as when training.
  In this case, in the direction of weak contact, $\bm{s}^{predict}_{seq}$ becomes $\bm{0}$ and the loss does not increase above $||\bm{0}, \bm{s}^{keep}_{seq}||^{2}_{2}$.
  On the other hand, in the direction of strong contact, the loss can increase steadily.
  As a result of optimization, $\bm{s}^{predict}_{seq}$ is likely to be $\bm{0}$.

  Therefore, in this study, we define the loss function as below,
  \begin{align}
    L_{grasp}(\bm{s}^{predict}_{seq}, \bm{s}^{keep}_{seq}) &\defeq \frac{1}{T}\sum^{T}_{i=1} w_{i}||\bm{s}^{predict}_{t+i} - \bm{s}^{keep}_{t+i}||^{2}_{2} \label{eq:good-loss}\\
    w_{i} &\defeq \begin{cases}
      1.0 & (\bm{s}^{predict}_{t+i} \geq \bm{s}^{keep}_{t+i})\nonumber\\
      C_{loss} & (otherwise)\nonumber
    \end{cases}
  \end{align}
  where $C_{loss}(C_{loss}>1)$ is a gain to increase loss when $\bm{s}^{predict}_{seq} \geq \bm{s}^{keep}_{seq}$.
  By using this $L_{grasp}$, grasping is stabilized while keeping the necessary contact.
  In this study, we set $C_{loss}=10.0$.

  To examine the correlation between the grasping stability and $L_{grasp}$, we conducted a simple experiment.
  We define $L_{grasp}$, considering only the current contact state $\bm{s}^{current}_{t}$, as $L_{eval}$, as shown below,
  \begin{align}
    L_{eval} &\defeq w_{0}||\bm{s}^{current}_{t} - \bm{s}^{keep}_{t}||^{2}_{2}\\
    w_{0} &\defeq \begin{cases}
      1.0 & (\bm{s}^{current}_{t} \geq \bm{s}^{keep}_{t})\nonumber\\
      C_{loss} & (otherwise)\nonumber
    \end{cases}
  \end{align}
  We conducted an experiment where Musashi grasps a hammer and hits a wooden plate successively, and obtained the sound value, which can be calculated by Fourier transform of the sound and then by extracting the maximum amplitude among a specific band, and $L_{eval}$.
  We show the three transitions of the sound value and $L_{eval}$ in \figref{figure:loss-sound}.
  $\sigma_{sound}$ expresses the variance of sound value.
  In the graphs from the left to right, $\sigma_{sound}$ and the change in $L_{eval}$ become larger.
  The sound value changes according to the changes in the contact state.
  We can see that $L_{eval}$ has a correlation with grasping stability.
  In this study, we optimize grasping stability based on $L_{grasp}$.
}%
{%
  最適化\equref{eq:opt}の損失関数$L_{opt}$の設計について説明する.
  なお, ここでは制御入力$\bm{u}$に関する制約については無視し, \secref{subsec:grasping-stabilizer}にて議論する.
  本研究では把持を安定化する, つまり, 現状の接触状態を維持し続けることを目的としている.
  ここで, 本研究で使用するような接触センサや筋張力センサの性質を考慮する必要がある.
  これらのセンサは接触がなくなると値が0になるのに対して, 接触が強くなる方向には定格まで常に値が大きくなり続ける.

  ここで, 損失関数$L_{opt}$を訓練時と同様に$L_{origin}$として設定したとする.
  このとき, $\bm{s}^{predict}_{seq}$が小さくなる方向に関しては途中でその値は$\bm{0}$となり, 損失は\textrm{MSE}$(\bm{0}, \bm{s}^{target}_{seq})$より大きくならない.
  しかし, $\bm{s}^{predict}_{seq}$が大きくなる方向は際限がない.
  ゆえに, 最適化の結果として$\bm{s}^{predict}_{seq}$は$\bm{0}$の方向に動きやすくなってしまう.

  そこで, 本研究では以下のように損失関数を定義する.
  \begin{align}
    L_{grasp}(\bm{s}^{predict}_{seq}, \bm{s}^{keep}_{seq}) &\defeq \frac{1}{T}\sum^{T}_{i=1} w_{i}||\bm{s}^{predict}_{t+i} - \bm{s}^{keep}_{t+i}||^{2}_{2} \label{eq:good-loss}\\
    w_{i} &\defeq \begin{cases}
      1.0 & (\bm{s}^{predict}_{t+i} \geq \bm{s}^{keep}_{t+i})\nonumber\\
      C_{loss} & (otherwise)\nonumber
    \end{cases}
  \end{align}
  ここで, $C_{loss}(C_{loss}>1)$は$\bm{s}^{predict}_{seq}$が$\bm{s}^{keep}_{seq}$より下がる方向に対しては損失を大きくするための係数を表す.
  この損失関数$L_{grasp}$を用いることで, 必要な接触を保ちつつ把持を安定化することができる.
  本研究では$C_{loss}=10.0$としている.

  この$L_{grasp}$と把持の安定性の相関を調べるために, 簡単な実験を行った.
  先ほどの$L_{grasp}$を現在の$\bm{s}^{current}_{t}$のみについて考えたものを以下のように$L_{eval}$として定義する.
  \begin{align}
    L_{eval} \defeq w_{0}||\bm{s}^{current}_{t} - \bm{s}^{keep}_{t}||^{2}_{2}
  \end{align}
  Musashiがハンマーを握って木の板を連続して叩き, その際の音をフーリエ変換して特定の周波数域の音の振幅の最大値を得る.
  この音の値とその際の$L_{eval}$の値の遷移の様子を3つ\figref{figure:loss-sound}に示す.
  $\sigma_{sound}$はsound valueの分散を表す.
  左から右に行くに従い, $L_{eval}$の値の変化が大きくなっていることがわかる.
  また, それに伴い$\sigma_{sound}$の分散も大きくなっている.
  $L_{eval}$は把持の安定性に相関があり, 把持状態が変化することで音の様子も変化してくことがわかる.
  よって, $L_{grasp}$をベースとして把持安定性を最適化していく.
}%

\subsection{Grasping Stabilizer} \label{subsec:grasping-stabilizer}
\switchlanguage%
{%
  We show the calculation procedures of \equref{eq:opt} as below.
  \begin{enumerate}
    \item Determine the initial control input $\bm{u}^{init}_{seq}$ before optimization
    \item Calculate the loss of $L_{opt}$
    \item Optimize $\bm{u}^{opt}_{seq}$ through backpropagation
  \end{enumerate}

  In 1), the determination of the initial control input is important since it largely affects the optimization result.
  In this study, we prepare a batch with $C_{const}+C_{opt}$ data including $C_{const}$ number of constant control inputs and $C_{opt}$ number of previously optimized results with noise.
  Regarding the former, the value from $\bm{u}^{min}$ to $\bm{u}^{max}$ is equally divided into $C_{opt}$ parts, and $\bm{u}^{init}_{[t, t+T-1]}$ filled with each value are obtained.
  Regarding the latter, we represent the previously optimized control input as $\bm{u}^{previous}_{[t-1, t+T-2]}$, and obtain $C_{opt}$ number of $\bm{u}^{init}_{seq}$ by adding together $\bm{u}^{previous}_{\{t, t+1, \cdots, t+T-2, t+T-2\}}$, which shifts $\bm{u}^{previous}_{[t-1, t+T-2]}$ and replicates the last term, and uniform random noise in the range of $[-0.1, 0.1]$.
  At $t=0$, the previously optimized value cannot be obtained, and so we fill the $\bm{u}^{previous}_{[t-1, t+T-2]}$ with $\bm{0}$.
  By starting from these initial control inputs, the optimized value with minimum $L_{opt}$ is sent to the actual robot.

  In 2), $L_{opt}$ is calculated as below,
  \begin{align}
    L_{opt} \defeq L_{grasp}(\bm{s}^{predict}_{seq}, \bm{s}^{keep}_{seq}) + C_{min}||\bm{u}^{init}_{seq}||^{2}_{2}\nonumber\\\;\;\;\;+ C_{adj}||\bm{u}^{init}_{[t, t+T-2]}-\bm{u}^{init}_{[t+1, t+T-1]}||^{2}_{2} \label{eq:opt-loss}
  \end{align}
  where $C_{min}$ and $C_{adj}$ are constant values.
  The second term of the right side of \equref{eq:opt-loss} is for minimizing the absolute value of control input, and the third term is for smoothing the transition of control input.
  $L_{opt}$ is calculated for each data in the batch of $\bm{u}^{init}_{seq}$, and the control input with minimum loss is defined as $\bm{u}^{opt}_{seq}$.

  In 3), $\bm{u}^{opt}_{seq}$ is optimized as below,
  \begin{align}
    \bm{g} &\defeq dL_{opt}/d\bm{u}^{opt}_{seq}\\
    \bm{u}^{opt}_{seq} &\gets \bm{u}^{opt}_{seq}-\gamma\bm{g}/||\bm{g}||_{2}
  \end{align}
  Since $\bm{u}^{opt}_{t+1}$ is sent to the actual robot as explained in \secref{subsec:formulation}, $\bm{u}^{opt}_{t}$ sent previously is not updated.
  $\gamma$ can be a constant value, but in this study, the best $\gamma$ is chosen by making a batch with various $\gamma$.
  The maximum value of $\gamma$, $\gamma_{max}$ is determined, the value from 0 to $\gamma_{max}$ is divided into $C^{opt}_{batch}$ parts, and a batch with $C^{opt}_{batch}$ number of data including $\bm{u}^{opt}_{seq}$ updated by each $\gamma$ is made.
  \equref{eq:opt-loss} is conducted again by setting $\bm{u}^{init}_{seq}\gets\bm{u}^{opt}_{seq}$, and $\bm{u}^{opt}_{seq}$ with minimum loss is adopted.
  By repeating the procedures of 2) and 3) $C^{opt}_{epoch}$ times, $\bm{u}^{opt}_{seq}$ is gradually optimized.

  $\bm{u}^{opt}_{t+1}$ in the finally obtained $\bm{u}^{opt}_{seq}$ is sent to the robot, and the grasp is stabilized.

  In this study, we set $C_{opt}=13$, $C_{const}=13$, $C_{min}=0.1$, $C_{adj}=0.1$, $\gamma_{max}=0.5$, $C^{opt}_{batch}=13$, and $C^{opt}_{epoch}=10$.
}%
{%
  \equref{eq:opt}の計算手順を以下に示す.
  \begin{enumerate}
    \item 最適化の初期値$\bm{u}^{init}_{seq}$の決定
    \item 損失関数$L_{opt}$の計算
    \item $\bm{u}^{opt}_{seq}$の最適化
  \end{enumerate}

  1)について, 最適化の結果は初期値に依存するため非常に重要である.
  本研究では一定制御入力の初期値$C_{const}$個と, 前回最適化結果にノイズを加えた値$C_{opt}$を足し合わせた$C_{const}+C_{opt}$個のデータをバッチとして用意する.
  前者は, $\bm{u}^{min}$から$\bm{u}^{max}$までを$C_{opt}$等分し, それぞれの値によって$[t, t+T-1]$個の時系列を埋めたデータである.
  また後者は, 前回最適化された$\bm{u}$を$\bm{u}^{previous}_{[t-1, t+T-2]}$として, この値をずらして最終項を複製した$\bm{u}^{previous}_{\{t, t+1, \cdots, t+T-2, t+T-2\}}$に[-0.1, 0.1]の一様なノイズを加えたデータを$C_{opt}$個用意する.
  $t=0$については前回最適化の結果がないため, 0で埋めた値に対してノイズを加えて用いている.
  これらの初期値から始め, 最終的に最も損失$L_{opt}$が小さくなったものを実機に送ることになる.

  次に, 2)の損失$L_{opt}$は以下のように計算する.
  \begin{align}
    L_{opt} \defeq L_{grasp}(\bm{s}^{predict}_{seq}, \bm{s}^{keep}_{seq}) + C_{min}||\bm{u}^{init}_{seq}||^{2}_{2}\nonumber\\\;\;\;\;+ C_{adj}||\bm{u}^{init}_{[t, t+T-2]}-\bm{u}^{init}_{[t+1, t+T-1]}||^{2}_{2} \label{eq:opt-loss}
  \end{align}
  ここで, $C_{min}, C_{adj}$は定数である.
  \equref{eq:opt-loss}の右辺第二項は制御入力をなるべく小さくする項であり, 第三項はタイムステップ間の制御入力の差を小さくし滑らかにする項である.
  この損失関数$L_{opt}$を$\bm{u}^{init}_{seq}$のバッチ内のデータそれぞれに対して計算し, 最も損失が小さかった値を$\bm{u}^{opt}_{seq}$とする.

  最後に, 3)では$\bm{u}^{opt}_{seq}$を以下のように最適化していく.
  \begin{align}
    \bm{g} &\defeq dL_{opt}/d\bm{u}^{opt}_{seq}\\
    \bm{u}^{opt}_{seq} &\gets \bm{u}^{opt}_{seq}-\gamma\bm{g}/||\bm{g}||_{2}
  \end{align}
  ここで, \secref{subsec:formulation}に述べたように$\bm{u}^{opt}_{t+1}$が実機に送られ, $\bm{u}^{opt}_{t}$は前回送られた固定の値であるため更新しない.
  このとき, $\gamma$の値を決め打ちしても良いが, 本研究では様々な$\gamma$によってバッチを作成し, 最も良い$\gamma$を選ぶ.
  $\gamma$の最大値$\gamma_{max}$を決め, 0から$\gamma_{max}$までの値を$C^{opt}_{batch}$等分し, それらそれぞれによって更新された$\bm{u}^{opt}_{seq}$を$C^{opt}_{batch}$個作成する.
  もう一度$\bm{u}^{init}_{seq}\gets\bm{u}^{opt}_{seq}$として2)の\equref{eq:opt-loss}を行い, 最も$L$が小さかった$\bm{u}^{opt}_{seq}$を採用する.
  上記2)--3)の工程を, $C^{opt}_{epoch}$回繰り返して$\bm{u}^{opt}_{seq}$を最適化していく.

  最終的に得られた$\bm{u}^{opt}_{seq}$における$\bm{u}^{opt}_{t+1}$を実機に送り, 把持を安定化する.

  本研究では, $C_{opt}=13$, $C_{const}=13$, $C_{min}=0.1$, $C_{adj}=0.1$, $\gamma_{max}=0.5$, $C^{opt}_{batch}=13$, $C^{opt}_{epoch}=10$としている.
}%

\begin{figure}[t]
  \centering
  \includegraphics[width=0.8\columnwidth]{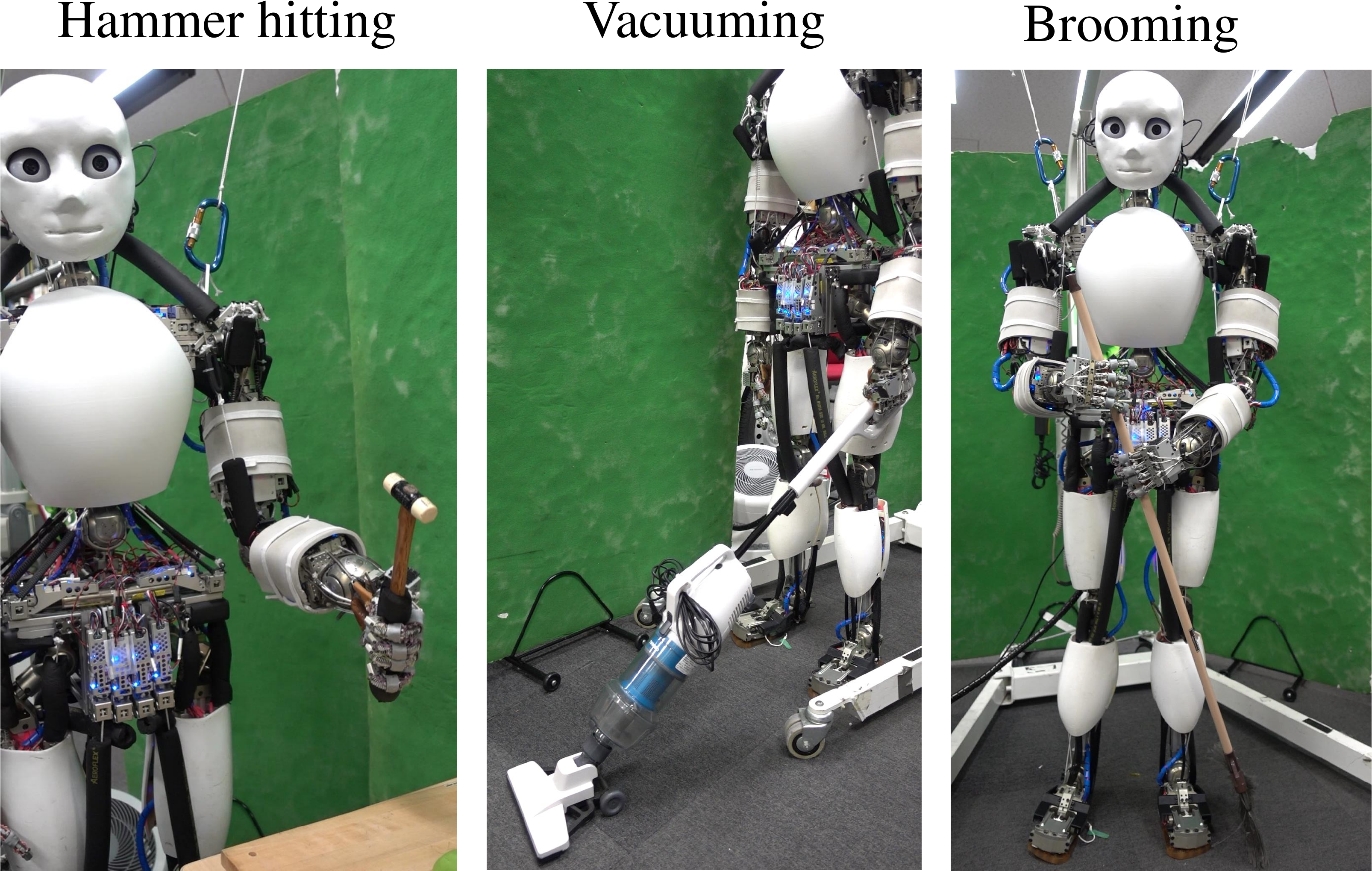}
  \caption{Experiments of hammer hitting, vacuuming, and brooming.}
  \label{figure:experimental-setup}
  \vspace{-1.0ex}
\end{figure}

\section{Experiments} \label{sec:experiments}
\switchlanguage%
{%
  We will verify the effectiveness of this study by experiments of hammer hitting, vacuuming, and brooming.
  Regarding the hammer hitting experiment, we will explain the search behavior, training results, and grasping stabilizer, in detail.

  We show each experiment in \figref{figure:experimental-setup}.
  Each tool is grasped by the left hand of Musashi, and each task is executed by swinging it.
  Regarding the brooming experiment, the grasping stabilizer is applied to only the left hand, and the broom is restrained into position at the right hand.
}%
{%
  ハンマー打撃動作, 箒掃き動作, 掃除動作を行うことで, 本研究の有効性を確認する.
  特にハンマー打撃動作については把持探索動作・訓練結果・把持安定化に分けて詳しく説明する.

  それぞれの実験の様子を\figref{figure:experimental-setup}に示す.
  基本的には左手で道具を把持し, それを振ることでタスクを実行する.
  箒については, 左手にのみGrasping Stabilizerを適用し, 右手には固定具を使用している.
}%

\begin{figure}[t]
  \centering
  \includegraphics[width=0.9\columnwidth]{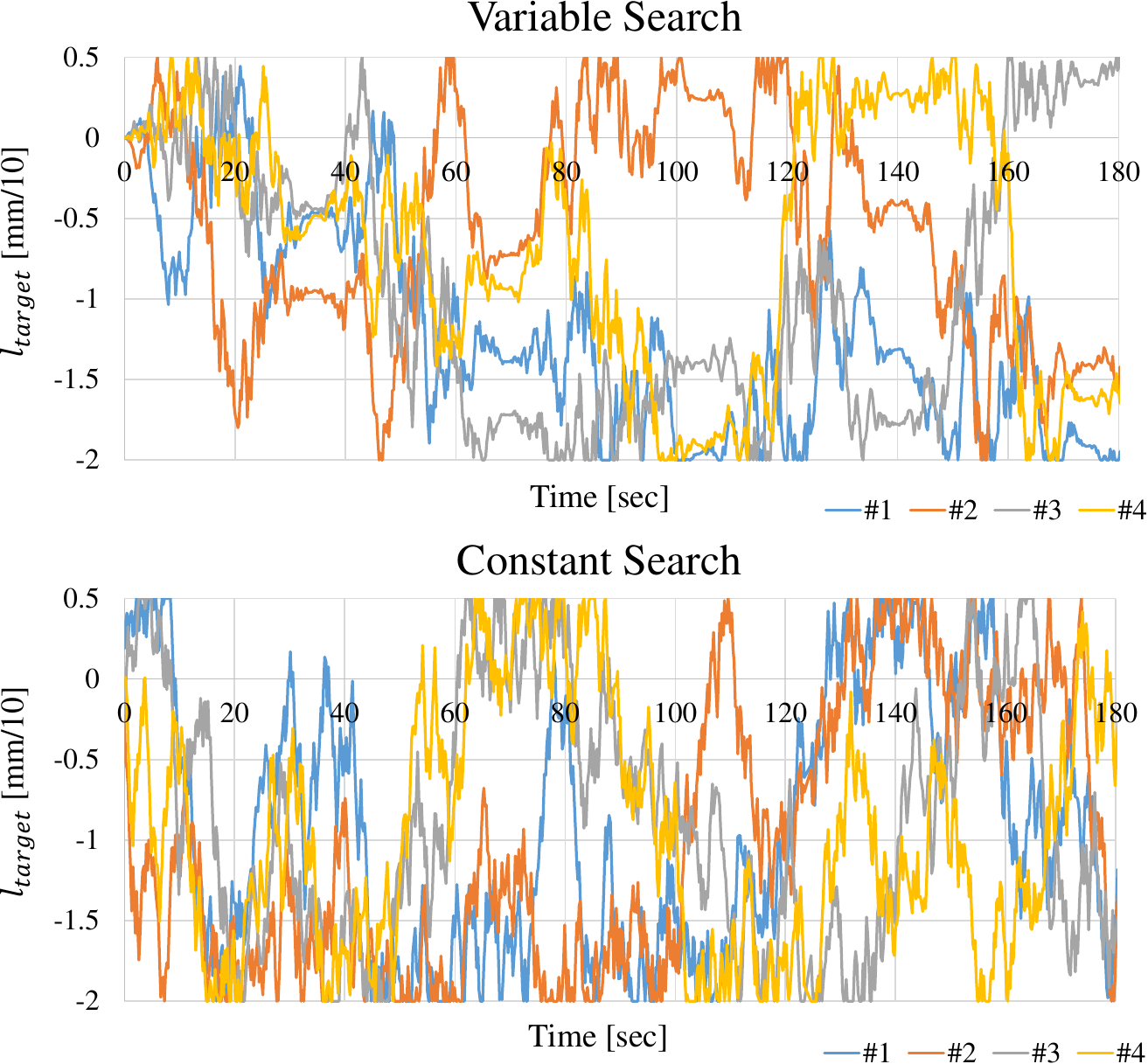}
  \caption{Transition of $\bm{u}$ during random search behaviors: Variable and Constant Search.}
  \label{figure:search-experiment}
  \vspace{-3.0ex}
\end{figure}

\begin{figure}[t]
  \centering
  \includegraphics[width=0.8\columnwidth]{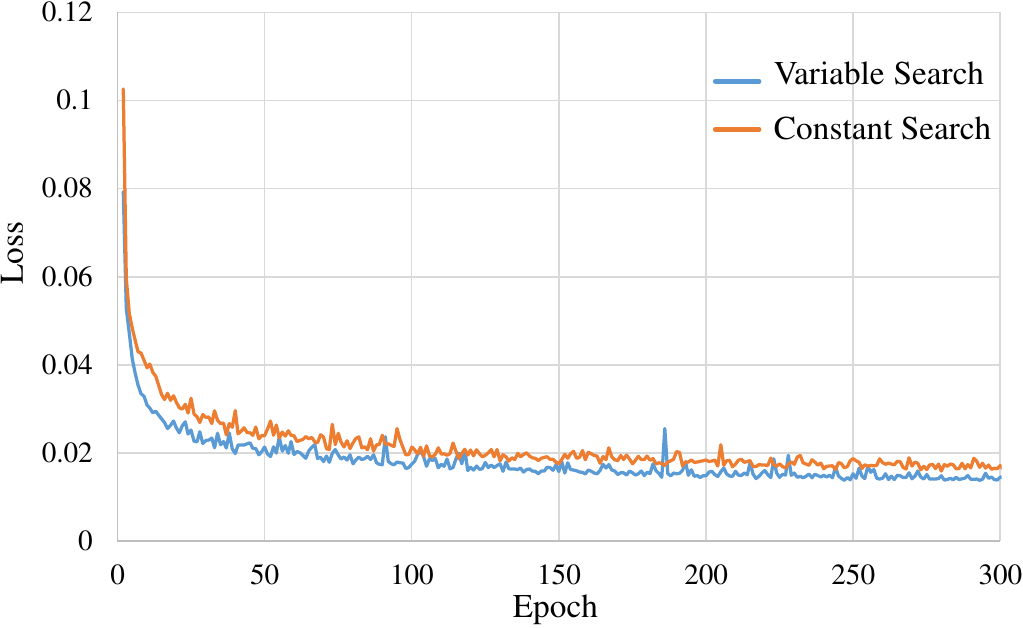}
  \caption{Transition of training loss $L_{origin}$ with data obtained using Variable or Constant Search.}
  \label{figure:train-experiment}
  \vspace{-3.0ex}
\end{figure}

\begin{figure*}[t]
  \centering
  \includegraphics[width=2.0\columnwidth]{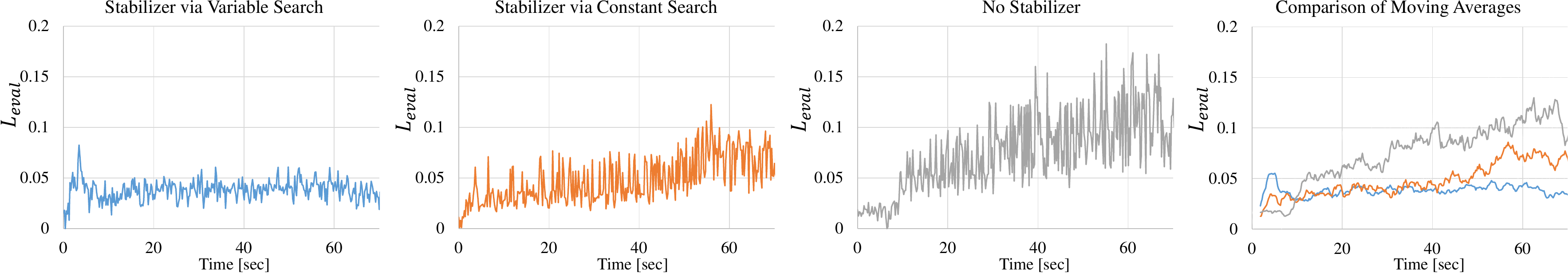}
  \caption{Hammer hitting experiment: comparison of transition of $L_{eval}$ among a stabilizer via Variable Search, a stabilizer via Constant Search, and no stabilizer}
  \label{figure:hammer-experiment}
  \vspace{-1.0ex}
\end{figure*}

\begin{figure*}[t]
  \centering
  \includegraphics[width=2.0\columnwidth]{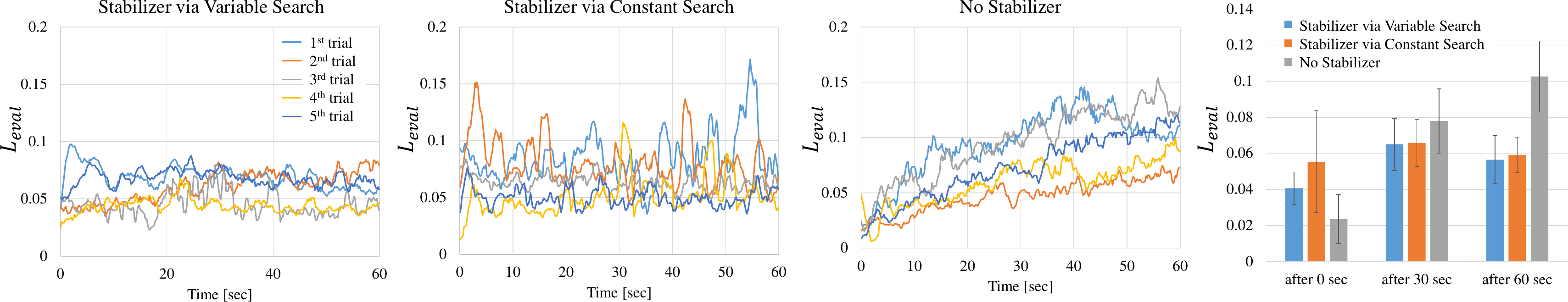}
  \caption{Quantitative evaluation of moving average of $L_{eval}$ after 0, 30, and 60 sec of hammer hitting with a stabilizer via Variable Search, a stabilizer via Constant Search, and without no stabilizer.}
  \label{figure:hammer-eval}
  \vspace{-1.0ex}
\end{figure*}

\subsection{Hammer Hitting} \label{subsec:hitting-experiment}
\subsubsection{Random Search Behavior and Training Phase}
\switchlanguage%
{%
  First, by randomly moving fingers, the actual robot sensor information for training of $\bm{f}$ can be obtained.
  We call a random search behavior with variable change in $\Delta{\bm{u}}$ explained in \secref{subsec:grasping-search} Variable Search, and an ordinary random walk with constant $\Delta{\bm{u}}$ (we set it as $C_{rand}$) Constant Search.
  The transition of $\bm{u}$ when grasping a hammer and conducting Variable or Constant Search is shown in \figref{figure:search-experiment}.
  Here, \#1 -- \#4 are the muscle numbers shown in the right figure of \figref{figure:musculoskeletal-hand}.
  While $\bm{u}$ vibrates at all times in Constant Search, there are sections where $\bm{u}$ vibrates and where $\bm{u}$ is smooth in Variable Search.
  900 numbers of data were obtained from the random search over 3 minutes.

  We trained the network $\bm{f}$ by splitting the obtained data into train and test (8:2).
  We show the transition of $L_{origin}$ regarding test data in \figref{figure:train-experiment}.
  The loss transitions with data when conducting Variable or Constant Search are almost the same, and the loss with smoother data when using Variable Search is slightly less than when using Constant Search.
}%
{%
  まず, ランダムに指を動かすことによって, 予測モデル学習のための実機データを取得する.
  ここで, \secref{subsec:grasping-search}に示したように, $\Delta{\bm{u}}$の値を変化させることによって実機データを学習させる方法をProposed Search, 通常のランダムウォークのように変化させずに一定値$C_{rand}$を用いる方法をConstant Searchと呼ぶこととする.
  このProposed/Constant Searchにおける$\bm{u}$の動きを\figref{figure:search-experiment}に示す.
  ここで, \#1 -- \#4は\figref{figure:musculoskeletal-hand}の右図におけるアクチュエータ番号と同等である.
  Constant Searchでは$\bm{u}$が常に激しく振動しているのにたいして, Propose Searchでは激しく振動する区間と滑らかな区間が存在していることがわかる.

  得られたデータを8:2でtrainとtest用に分けて学習し, testデータに関する損失関数$L_{origin}$の遷移を\figref{figure:train-experiment}に示す.
  概ね同じような遷移をしており, Variable Searchの方が滑らかなデータが増えるため損失は少しだけ少ないことがわかる.
}%

\subsubsection{Grasping Stabilizer}
\switchlanguage%
{%
  We conducted experiments of grasping stabilizer using the trained model $\bm{f}$.
  The robot grasped a hammer and continued to hit at certain intervals.
  We show the comparison of the transition of $L_{eval}$ when using a stabilizer via Variable Search, a stabilizer via Constant Search, and no stabilizer, in \figref{figure:hammer-experiment}.
  Respective transitions are shown in the three left graphs, and the right graph shows the comparison of the transition of each moving average over 2sec.
  Compared to no stabilizer, by using stabilizers, $L_{eval}$ is small and the initial contact state can be kept.
  When using a stabilizer via Variable Search, the average of $L_{eval}$ and its variance are smaller than via Constant Search.
  This is because the obtained $\bm{u}$ in Constant Search vibrates at all times and so the result of optimization using the vibrated data also vibrates.

  Also, we repeated the same hammer experiment 5 times.
  We show the transition of each moving average of $L_{eval}$ over 2 sec, and the average and variance of the value after hitting movements over 0, 30, and 60 sec, in \figref{figure:hammer-eval}.
  We can see the same tendency in five trials from the three left graphs.
  Regarding the average of $L_{eval}$ after 0 sec, no stabilizer is the best and a stabilizer via Constant Search is the worst.
  This is because control input using a stabilizer with Constant Search vibrates and rapidly breaks the initial contact state, while no stabilizer can keep the initial contact state at the initial stage of tool-use.
  However, after tool-use over 60 sec, a stabilizer with Variable Search is the best, and the initial contact state gradually changes without stabilizer.
}%
{%
  得られた予測モデル$\bm{f}$を用いて把持安定化を行う.
  ハンマーを持ち, 一定間隔で叩き続ける.
  この際の$L_{eval}$の遷移について, $\bm{f}$をPropose Searchのデータによって学習して把持最適化に用いた場合, $\bm{f}$をConstant Searchのデータによって学習して把持最適化に用いた場合, 一切の把持安定化を行わない場合で比較実験をする.
  その結果を\figref{figure:hammer-experiment}に示す.
  一番右の図は, 2 sec間の移動平均を取った比較図となる.
  把持安定化をしない場合に比べて, 安定化をした場合のほうが$L_{eval}$は小さくなり, 初期の接触を保ち続けられていることがわかる.
  予測モデルの学習にProposed Searchを使った場合は, Constant Searchを使った場合に比べて$L_{eval}$の平均が小さく, 分散も小さいことがわかる.
  これは, $\bm{u}$について, 振動しがちなデータのみを与えてしまったため, 実際の最適化も振動しがちになってしまったということが考えられる.

  また, これら3つについて5回ずつ実験を行い, $L_{eval}$の2 sec間の移動平均に関して, 動作開始から0, 30, 60秒後の値の平均と分散を計算し, \figref{figure:hammer-eval}に示す.
  \figref{figure:hammer-eval}の左3つはそれぞれに関する移動平均であるが, 5回全てに同じような傾向が見られる.
  0 secにおける平均は最適化しないときが最も良く, constant searchを使った場合が最も悪い.
  これは, 道具使用初期は何も動作しない方が初期状態を保つことができるのに対して, constant searchを使ったstabilizerでは動作が振動してしまいすぐに接触状態を崩してしまっていることが原因であると考える.
  しかし, 60 sec後はstabilizer with Variable Searchが最も良く, stabilizerを入れない方法では徐々に接触状態が変化してしまっていることがわかる.
}%

\begin{figure}[t]
  \centering
  \includegraphics[width=0.9\columnwidth]{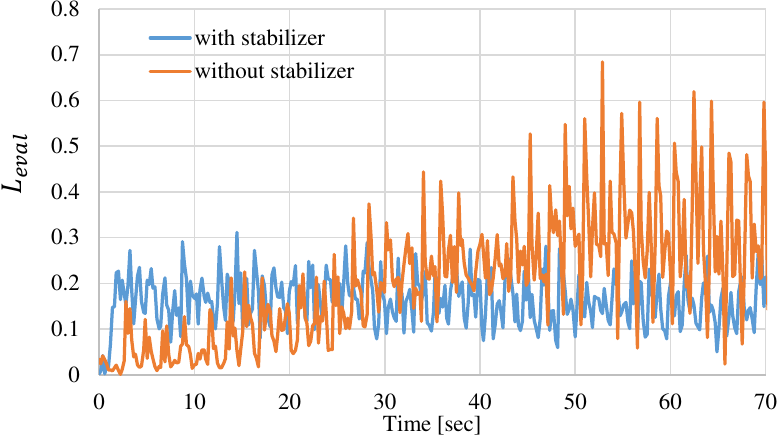}
  \caption{Vacuuming experiment: comparison of transition of $L_{eval}$ between with and without stabilizer.}
  \label{figure:vacuuming-experiment}
  \vspace{-3.0ex}
\end{figure}

\subsection{Vacuuming} \label{subsec:vacuuming-experiment}
\switchlanguage%
{%
  We conducted a vacuuming experiment.
  In the following sections, we simply refer to a stabilizer via variable search as ``a stabilizer''.
  As in \secref{subsec:hitting-experiment}, the model was trained via Variable Search when grasping a vacuum cleaner, and the grasping stabilizer was executed.
  We show the result in \figref{figure:vacuuming-experiment}.
  After tool-use without stabilizer over 70 sec, the contact state changes and $L_{eval}$ largely vibrates by impacts from the tool.
  On the other hand, $L_{eval}$ is kept constant with a stabilizer.
}%
{%
  掃除機を前後に動かす動作を行う.
  以降では stabilizer with Variable Searchを単純にstabilizerと呼び使用する.
  70 secの動作では, stabilizerを入れない場合は徐々に接触状態が変化し, かつ衝撃によって大きく振動している.
  それに対して, stabilizerを入れた場合は$L_{eval}$が一定値を保っていることがわかる.
}%

\begin{figure}[t]
  \centering
  \includegraphics[width=0.9\columnwidth]{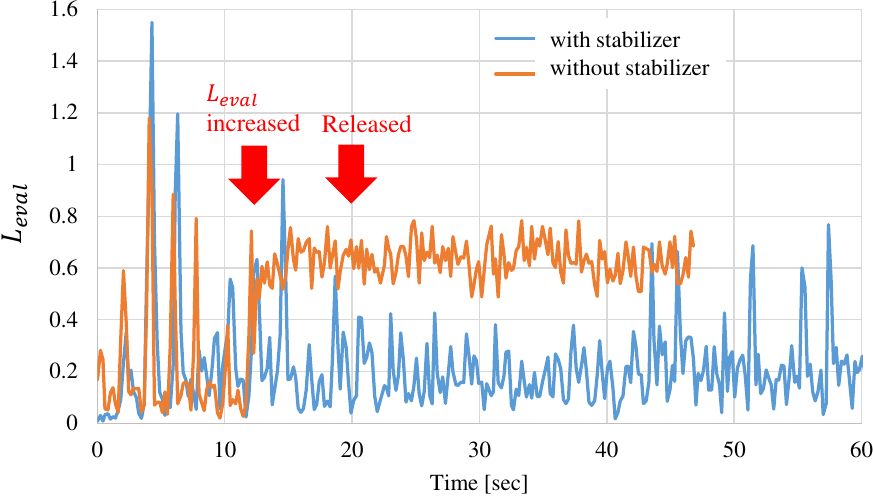}
  \caption{Brooming experiment: comparison of transition of $L_{eval}$ between with and without stabilizer.}
  \label{figure:brooming-experiment}
  \vspace{-1.0ex}
\end{figure}

\begin{figure}[t]
  \centering
  \includegraphics[width=0.8\columnwidth]{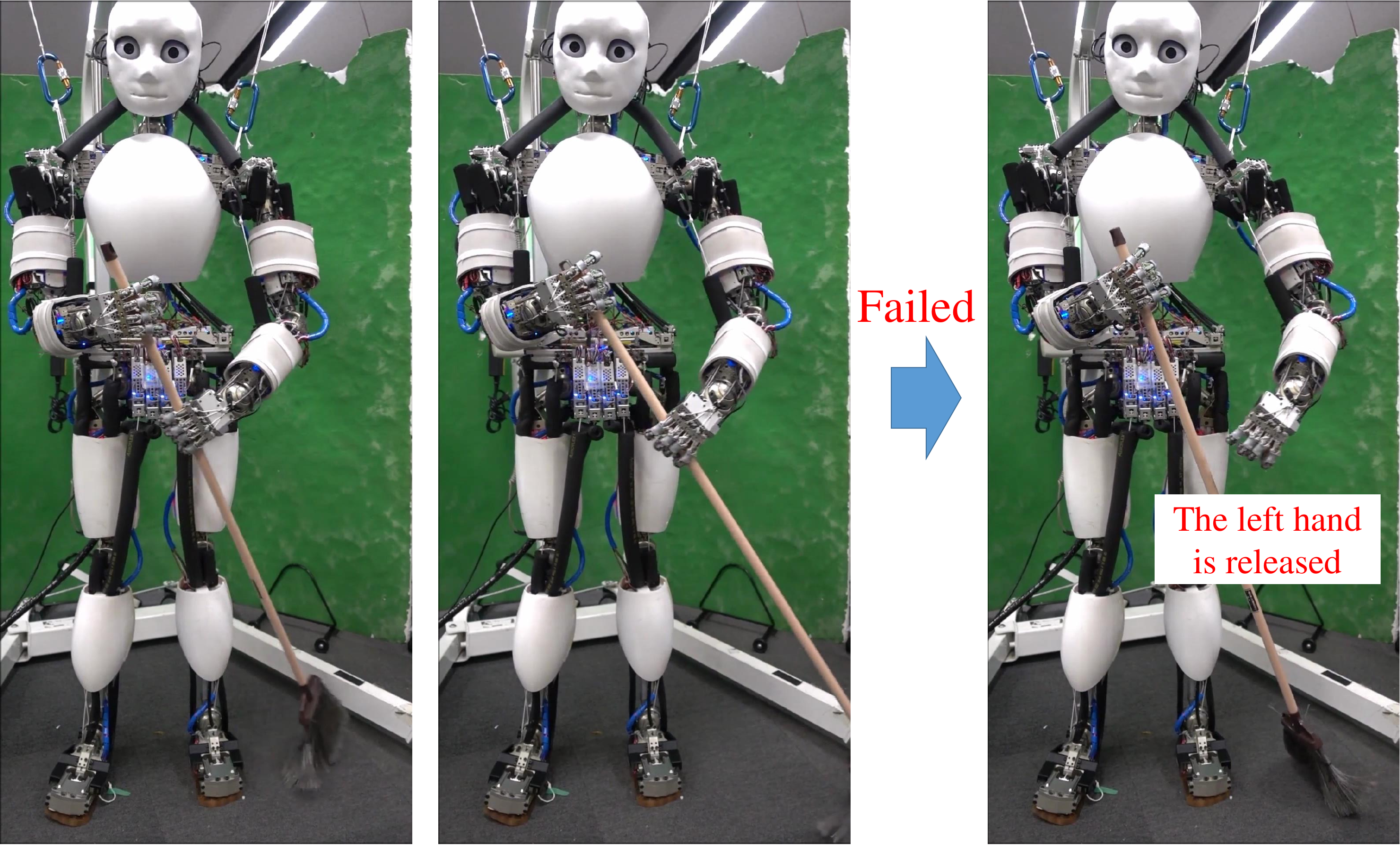}
  \caption{The failure of brooming.}
  \label{figure:brooming-failure}
  \vspace{-3.0ex}
\end{figure}

\subsection{Brooming} \label{subsec:brooming-experiment}
\switchlanguage%
{%
  We conducted a brooming experiment by Musashi.
  As in \secref{subsec:hitting-experiment}, the model was trained via Variable Search when grasping a broomstick, and the grasping stabilizer was executed.
  We show the transition of $L_{eval}$ in \figref{figure:brooming-experiment}, regarding with and without stabilizer.
  Both $L_{eval}$ with and without stabilizer vibrated largely at the initial stage of tool-use, but the value with stabilizer could be kept constant.
  On the other hand, $L_{eval}$ without stabilizer rose largely after 12 sec, and the brooming failed after 20 sec.
  The failure of brooming is shown in \figref{figure:brooming-failure}, and the left hand was released from the broomstick.

  Also, we conducted the brooming experiment of 60 sec 5 times with and without stabilizer, and show the success (\checkmark) or failure (the time of failure) of each trial in \tabref{table:brooming-experiment}.
  With stabilizer, brooming succeeded 3 / 5 times over 60 sec, and the experiments without stabilizer failed after average of 28.6 sec.
}%
{%
  Musashiによって箒を掃く実験を行う.
  その際の$L_{eval}$の遷移を\figref{figure:brooming-experiment}に示す.
  stabilizerを入れた場合, 入れない場合に関して両者とも初期に大きく$L_{eval}$がズレているが, stabilizerを入れた場合はその後一定の値に収束している.
  それに対して, stabilizerを入れない場合は12 secで$L_{eval}$が大きく上昇し, 20 secで箒掃きが失敗している.
  その際の様子が\figref{figure:brooming-failure}であり, 左手が箒から離れてしまったのがわかる.

  また, この箒掃き実験をstabilizerを入れた場合と入れない場合に関して60秒間を5回ずつ行い, その際の成功・失敗と, 失敗した場合の秒数を\tabref{table:brooming-experiment}に示す.
  stabilizerを入れた場合は5回中3回は60秒間の動きを成功させ, 入れない場合は平均28.6秒で箒から手が離れてしまった.
}%

\begin{table}[htb]
  \centering
  \caption{Quantitative evaluation of the brooming experiment. {\checkmark}s express that the experiment succeeded over 60 sec, and the numbers expresse the time of failure.}
  \begin{tabular}{l|ccccc}
    Number of trials & $1^{st}$ & $2^{nd}$ & $3^{rd}$ & $4^{th}$ & $5^{th}$\\ \hline
    with stabilizer & \checkmark & 23 & \checkmark & \checkmark & 47\\
    without stabilizer & 20 & 33 & 4 & 52 & 34 \\
  \end{tabular}
  \label{table:brooming-experiment}
\end{table}

\section{Discussion} \label{sec:discussion}
\switchlanguage%
{%
  From the experiments, we can see that tool-use is stabilized using the grasping stabilizer proposed in this study.
  Regarding the hammer hitting and vacuuming experiments, although we cannot see a visual difference, the initial contact is kept using grasping stabilizer.
  Also, in the brooming experiment, the stabilizer can inhibit failure of releasing the hand from the broomstick by impacts and external force to the tool.

  However, several problems remain.
  First, the thumb sometimes moves to postures impossible for human beings due to the random search of muscle space.
  To solve the problem, we should add some constraints for the movements of muscles related to the thumb.
  Second, regarding the brooming task, the hand is sometimes released even if the grasping stabilizer is used.
  This is due to a strong impact from the friction with the floor, and so we need to collect the motion data of such situations and train the network using it.
  Third, the motions of the experiments are quite slow.
  To stabilize the grasping even when more dynamic movements are conducted, we need to speed up the control frequency.
  For the high frequency, we need to accelerate the optimization of this study using more calculation resources or simplifying the network structure.

  Because this study focuses not on the problem of searching a wide space as handled in reinforcement learning, but on stabilization of tool-use, we can apply this method to the actual robot only by training the contact state transition around the initial contact state.
  The grasping stabilizer can sufficiently work by random search behavior over just 3 minutes.
  Although we use 4 muscle lengths as control input in this study, in order to use more complex hands with many muscles, we may need to consider muscle synergy \cite{allessandro2013synergy}.

  The important point of this study is the implicit training of the nontrivial dynamic relationship between contact sensors and actuators of flexible hands.
  This study can be applied to various hands other than flexible musculoskeletal hands, and stable grasping and tool-use are expected.
}%
{%
  本研究の実験から, grasping stabilizerを入れることで, 道具把持を安定化することができることがわかった.
  ハンマー動作や掃除機使用においては, 視覚的な変化は見られなかったものの, stabilizerを入れることで初期接触状態を入れない場合に比べて維持することができることがわかった.
  また, 箒掃き動作からは, 外力によって手が離れてしまうような失敗を, stabilizerを入れることによって緩和することができることがわかった.

  しかし, いくつかの課題が残った.
  まず, 親指の動きは, ランダムに動作させることで, 通常人間ではあり得ないような把持・接触をしてしまう場合がある.
  これを解決するためには, 単純に筋をランダムに動かすのではなく, 親指に関わる筋のランダム性に制約を入れる必要があると考える.
  次に, 箒タスクについて, stabilizerを入れても手が離れてしまう場合があった.
  これは床との摩擦による強い衝撃に耐えられなかったのが原因と考えられるが, それらの状況も含めたランダム探索を行い, よりデータを増やして対応していく必要がある.
  最後に, 本研究では動作のスピードが遅いという問題がある.
  よりダイナミックに動作させるためには, より制御周期を速くする必要があると考える.
  そのためには, ネットワークの簡易化による最適化の高速化等の工夫が今後必要になると考える.

  本研究は, 強化学習等における非常に探索空間の広い問題ではなく, 把持安定化に着目することで, 初期把持における接触状態の近辺を探索して状態遷移方程式を学習するのみで実機に適用することができる.
  そのため, 3分程度の実機学習でも十分に効果を示すことができた.
  本研究では4本の筋長という制御入力を用いたが, より複雑なハンドでは探索空間が増え学習が難しくなるため, 筋シナジー等の考え方を導入する必要があるかもしれない.

  本研究で重要な点は, 柔軟ハンドにおける非自明な接触センサや筋長センサと制御入力の関係を暗に学習しているところである.
  これまで多く行われてきた, それらをモデルから計算できるようなハンド以外にも本研究は適用可能であり, 信頼性のある柔軟ハンドによる把持・道具操作等が期待される.
}%

\section{CONCLUSION} \label{sec:conclusion}
\switchlanguage%
{%
  In this study, we proposed a strategy for stable tool-use based on the construction of a predictive model of sensor state transition and optimization of control input.
  Regarding flexible under-actuated hands, since the relationship between sensors and actuators cannot be uniquely determined, we must train the predictive model and optimize time-series control input for the grasping stabilizer.
  By using backpropagation technique of a neural network for the control input and exploring random search behavior, design of loss function, and optimization method, tool grasping is stabilized.
  Especially, to obtain a better stabilizer, the random search method varying the motion speed is necessary.
  Also, the loss function should consider the anisotropy of contact sensor values in the positive and negative direction.
  The predictive model is sufficiently constructed over 3 minutes of search behavior, and a grasping stabilizer adapted to the tool can be obtained.

  In future works, we would like to apply this study to multiple tools by inputting tool images, and explore in-hand manipulation by flexible hands.
}%
{%
  本研究では, 接触状態遷移の予測モデル構築と最適化に基づく道具把持安定化戦略について述べた.
  柔軟で劣駆動なハンドではセンサとアクチュエータの関係が一意に決まらないため, 予測モデルを学習することで時系列に安定化のための制御入力を最適化することができる.
  ニューラルネットワークの誤差逆伝播を制御入力に対して用い, 探索方法や損失関数の設計, 最適化手法を工夫することで把持安定化が可能となる.
  特に, 良い安定化戦略を得るためには, ランダムサーチにおける動きに緩急をつけてあげると良い.
  また, 接触センサの正負への値変化の異方性を考慮して損失関数を設計してあげる必要がある.
  予測モデルは3分程度の探索行動で十分に構築することができ, その道具に適応した把持安定化制御を得ることが可能である.

  今後は, 道具の画像を入力に加えより広く道具操作に活用できること, In-handマニピュレーションへの発展等が考えられる.
}%

{
  %\footnotesize
  %\small
  %\bibliographystyle{junsrt}
  \bibliographystyle{IEEEtran}
  \bibliography{main}

\begin{thebibliography}{10}
\providecommand{\url}[1]{#1}
\csname url@rmstyle\endcsname
\providecommand{\newblock}{\relax}
\providecommand{\bibinfo}[2]{#2}
\providecommand\BIBentrySTDinterwordspacing{\spaceskip=0pt\relax}
\providecommand\BIBentryALTinterwordstretchfactor{4}
\providecommand\BIBentryALTinterwordspacing{\spaceskip=\fontdimen2\font plus
\BIBentryALTinterwordstretchfactor\fontdimen3\font minus
  \fontdimen4\font\relax}
\providecommand\BIBforeignlanguage[2]{{%
\expandafter\ifx\csname l@#1\endcsname\relax
\typeout{** WARNING: IEEEtran.bst: No hyphenation pattern has been}%
\typeout{** loaded for the language `#1'. Using the pattern for}%
\typeout{** the default language instead.}%
\else
\language=\csname l@#1\endcsname
\fi
#2}}

\bibitem{kochan2005shadowhand}
A.~Kochan, ``{Shadow delivers first hand},'' \emph{Industrial Robot}, vol.~32,
  no.~1, pp. 15--16, 2005.

\bibitem{grebenstein2011dlrhand}
M.~Grebenstein, A.~Albu-Sch{\"a}ffer, T.~Bahls, M.~Chalon, O.~Eiberger,
  W.~Friedl, R.~Gruber, S.~Haddadin, U.~Hagn, R.~Haslinger, H.~H{\"o}ppner,
  S.~J{\"o}rg, M.~Nickl, A.~Nothhelfer, F.~Petit, J.~Reill, N.~Seitz,
  T.~Wimb{\"o}ck, S.~Wolf, T.~W{\"u}sthoff, and G.~Hirzinger, ``{The DLR hand
  arm system},'' in \emph{Proceedings of the 2011 IEEE International Conference
  on Robotics and Automation}, 2011, pp. 3175--3182.

\bibitem{kim2014roborayhand}
Y.~Kim, Y.~Lee, J.~Kim, J.~Lee, K.~Park, K.~Roh, and J.~Choi, ``{RoboRay hand:
  A highly backdrivable robotic hand with sensorless contact force
  measurements},'' in \emph{Proceedings of the 2014 IEEE International
  Conference on Robotics and Automation}, 2014, pp. 6712--6718.

\bibitem{deimel2016underactuatedhand}
R.~Deimel and O.~Brock, ``{A novel type of compliant and underactuated robotic
  hand for dexterous grasping},'' \emph{The International Journal of Robotics
  Research}, vol.~35, no. 1--3, pp. 161--185, 2016.

\bibitem{wiste2017anthrohand}
T.~Wiste and M.~Goldfarb, ``{Design of a simplified compliant anthropomorphic
  robot hand},'' in \emph{Proceedings of the 2017 IEEE International Conference
  on Robotics and Automation}, 2017, pp. 3433--3438.

\bibitem{xu2016biohand}
Z.~Xu and E.~Todorov, ``{Design of a highly biomimetic anthropomorphic robotic
  hand towards artificial limb regeneration},'' in \emph{Proceedings of the
  2016 IEEE International Conference on Robotics and Automation}, 2016, pp.
  3485--3492.

\bibitem{kontoudis2015lockablehand}
G.~P. Kontoudis, M.~V. Liarokapis, A.~G. Zisimatos, C.~I. Mavrogiannis, and
  K.~J. Kyriakopoulos, ``{Open-source, anthropomorphic, underactuated robot
  hands with a selectively lockable differential mechanism: Towards affordable
  prostheses},'' in \emph{Proceedings of the 2015 IEEE/RSJ International
  Conference on Intelligent Robots and Systems}, 2015, pp. 5857--5862.

\bibitem{makino2017hand}
S.~Makino, K.~Kawaharazuka, M.~Kawamura, Y.~Asano, K.~Okada, and M.~Inaba,
  ``{High-power, flexible, robust hand: Development of musculoskeletal hand
  using machined springs and realization of self-weight supporting motion with
  humanoid},'' in \emph{Proceedings of the 2017 IEEE/RSJ International
  Conference on Intelligent Robots and Systems}, 2017, pp. 1187--1192.

\bibitem{makino2018hand}
S.~Makino, K.~Kawaharazuka, M.~Kawamura, A.~Fujii, T.~Makabe, M.~Onitsuka,
  Y.~Asano, K.~Okada, K.~Kawasaki, and M.~Inaba, ``{Five-Fingered Hand with
  Wide Range of Thumb Using Combination of Machined Springs and Variable
  Stiffness Joints},'' in \emph{Proceedings of the 2018 IEEE/RSJ International
  Conference on Intelligent Robots and Systems}, 2018, pp. 4562--4567.

\bibitem{lee2017softrobotics}
C.~Lee, M.~Kim, Y.~J. Kim, N.~Hong, S.~Ryu, H.~J. Kim, and S.~Kim, ``{Soft
  robot review},'' \emph{International Journal of Control, Automation and
  Systems}, vol.~15, no.~1, pp. 3--15, 2017.

\bibitem{allen1997feedback}
P.~K. Allen, A.~T. Miller, P.~Y. Oh, and B.~S. Leibowitz, ``{Using tactile and
  visual sensing with a robotic hand},'' in \emph{Proceedings of the 1997 IEEE
  International Conference on Robotics and Automation}, 1997, pp. 676--681.

\bibitem{bicchi1989feedback}
A.~Bicchi, J.~K. Salisbury, and P.~Dario, ``{Augmentation of grasp robustness
  using intrinsic tactile sensing},'' in \emph{Proceedings of the 1989 IEEE
  International Conference on Robotics and Automation}, 1989, pp. 302--307.

\bibitem{regoli2016grasp}
M.~Regoli, U.~Pattacini, G.~Metta, and L.~Natale, ``{Hierarchical grasp
  controller using tactile feedback},'' in \emph{Proceedings of the 2016
  IEEE-RAS International Conference on Humanoid Robots}, 2016, pp. 387--394.

\bibitem{schmid2008door}
A.~J. Schmid, N.~Gorges, D.~Goger, and H.~Worn, ``{Opening a door with a
  humanoid robot using multi-sensory tactile feedback},'' in \emph{Proceedings
  of the 2008 IEEE International Conference on Robotics and Automation}, 2008,
  pp. 285--291.

\bibitem{hogan2018regrasp}
F.~R. Hogan, M.~Bauza, O.~Canal, E.~Donlon, and A.~Rodriguez, ``{Tactile
  Regrasp: Grasp Adjustments via Simulated Tactile Transformations},'' in
  \emph{Proceedings of the 2018 IEEE/RSJ International Conference on
  Intelligent Robots and Systems}, 2018, pp. 2963--2970.

\bibitem{calandra2018regrasp}
R.~Calandra, A.~Owens, D.~Jayaraman, J.~Lin, W.~Yuan, J.~Malik, E.~H. Adelson,
  and S.~Levine, ``{More Than a Feeling: Learning to Grasp and Regrasp Using
  Vision and Touch},'' \emph{IEEE Robotics and Automation Letters}, vol.~3,
  no.~4, pp. 3300--3307, 2018.

\bibitem{li2015garments}
Y.~Li, D.~Xu, Y.~Yue, Y.~Wang, S.~Chang, E.~Grinspun, and P.~K. Allen,
  ``{Regrasping and unfolding of garments using predictive thin shell
  modeling},'' in \emph{Proceedings of the 2015 IEEE International Conference
  on Robotics and Automation}, 2015, pp. 1382--1388.

\bibitem{chebotar2016regrasping}
Y.~Chebotar, K.~Hausman, Z.~Su, G.~S. Sukhatme, and S.~Schaal,
  ``{Self-supervised regrasping using spatio-temporal tactile features and
  reinforcement learning},'' in \emph{Proceedings of the 2016 IEEE/RSJ
  International Conference on Intelligent Robots and Systems}, 2016, pp.
  1960--1966.

\bibitem{jain2019manipulation}
D.~Jain, A.~Li, S.~Singhal, A.~Rajeswaran, V.~Kumar, and E.~Todorov,
  ``{Learning Deep Visuomotor Policies for Dexterous Hand Manipulation},'' in
  \emph{Proceedings of the 2019 IEEE International Conference on Robotics and
  Automation}, 2019, pp. 3636--3643.

\bibitem{hoof2015reinforcement}
H.~V. Hoof, T.~Hermans, G.~Neumann, and J.~Peters, ``{Learning robot in-hand
  manipulation with tactile features},'' in \emph{Proceedings of the 2015
  IEEE-RAS International Conference on Humanoid Robots}, 2015, pp. 121--127.

\bibitem{homberg2019soft}
B.~S. Homberg, R.~K. Katzschmann, M.~R. Dogar, and D.~Rus, ``{Robust
  proprioceptive grasping with a soft robot hand},'' \emph{Autonomous Robots},
  vol.~43, no.~3, pp. 681--696, 2019.

\bibitem{kawaharazuka2019musashi}
K.~Kawaharazuka, S.~Makino, K.~Tsuzuki, M.~Onitsuka, Y.~Nagamatsu, K.~Shinjo,
  T.~Makabe, Y.~Asano, K.~Okada, K.~Kawasaki, and M.~Inaba, ``{Component
  Modularized Design of Musculoskeletal Humanoid Platform Musashi to
  Investigate Learning Control Systems},'' in \emph{Proceedings of the 2019
  IEEE/RSJ International Conference on Intelligent Robots and Systems}, 2019,
  pp. 7294--7301.

\bibitem{kawaharazuka2019dynamic}
K.~Kawaharazuka, T.~Ogawa, J.~Tamura, and C.~Nabeshima, ``{Dynamic Manipulation
  of Flexible Objects with Torque Sequence Using a Deep Neural Network},'' in
  \emph{Proceedings of the 2019 IEEE International Conference on Robotics and
  Automation}, 2019, pp. 2139--2145.

\bibitem{kawaharazuka2019pedal}
K.~Kawaharazuka, K.~Tsuzuki, S.~Makino, M.~Onitsuka, K.~Shinjo, Y.~Asano,
  K.~Okada, K.~Kawasaki, and M.~Inaba, ``{Task-specific Self-body Controller
  Acquisition by Musculoskeletal Humanoids: Application to Pedal Control in
  Autonomous Driving},'' in \emph{Proceedings of the 2019 IEEE/RSJ
  International Conference on Intelligent Robots and Systems}, 2019, pp.
  813--818.

\bibitem{kawaharazuka2017forearm}
K.~Kawaharazuka, S.~Makino, M.~Kawamura, Y.~Asano, Y.~Kakiuchi, K.~Okada, and
  M.~Inaba, ``{Human Mimetic Forearm Design with Radioulnar Joint using
  Miniature Bone-muscle Modules and its Applications},'' in \emph{Proceedings
  of the 2017 IEEE/RSJ International Conference on Intelligent Robots and
  Systems}, 2017, pp. 4956--4962.

\bibitem{hochreiter1997lstm}
S.~Hochreiter and J.~Schmidhuber, ``{Long short-term memory},'' \emph{Neural
  computation}, vol.~9, no.~8, pp. 1735--1780, 1997.

\bibitem{ioffe2015batchnorm}
S.~Ioffe and C.~Szegedy, ``{Batch Normalization: Accelerating Deep Network
  Training by Reducing Internal Covariate Shift},'' in \emph{Proceedings of the
  32nd International Conference on Machine Learning}, 2015, pp. 448--456.

\bibitem{allessandro2013synergy}
C.~Alessandro, I.~Delis, F.~Nori, S.~Panzeri, and B.~Berret, ``{Muscle
  synergies in neuroscience and robotics: from input-space to task-space
  perspectives},'' \emph{Frontiers in Computational Neuroscience}, vol.~7,
  no.~43, pp. 1--16, 2013.

\end{thebibliography}
}

\end{document}